\documentclass[10pt,twocolumn,letterpaper]{article}

\usepackage{cvpr}
\usepackage{times}
\usepackage{epsfig}
\usepackage{graphicx}
\usepackage{amsmath}
\usepackage{amssymb}
\usepackage{times}
\usepackage{soul}
\usepackage{url}
\usepackage[utf8]{inputenc}

\usepackage{amsmath}
\usepackage{amsthm}
\usepackage{booktabs}
\usepackage{algorithm}
\usepackage{algorithmic}
\usepackage{subfigure}
\usepackage{amsmath}
\usepackage{amssymb}
\usepackage{booktabs}
\usepackage{multirow}
\usepackage{paralist,algorithmic,algorithm}
\usepackage[american]{babel}
\usepackage{microtype}
\usepackage{lipsum}
\usepackage{bm}
\usepackage[switch]{lineno}  %
\DeclareMathOperator*{\argmax}{argmax} 
\newcommand{\x}{{\bf x}}
\newcommand{\w}{{\bm w}}

\newcommand{\D}{\mathcal{D}}

\newcommand{\firstcite}[1]{\cite{#1}}

\newcommand{\bfname}[1]{{\bf #1}}
\newcommand{\name}{{\sc Proser }}
\newcommand{\mame}{{\sc Proser}}
\newcommand{\argmin}{\mathop{\arg\!\min}}
\newcommand{\nrr}{N. R.}

\usepackage[pagebackref=true,breaklinks=true,letterpaper=true,colorlinks,bookmarks=false]{hyperref}

\cvprfinalcopy % *** Uncomment this line for the final submission

\def\cvprPaperID{4081} % *** Enter the CVPR Paper ID here

\ifcvprfinal\fi
\begin{document}

%%%%%%%%% TITLE
\title{Learning Placeholders for Open-Set Recognition}

\author{Da-Wei Zhou\quad \qquad Han-Jia Ye\footnotemark[2] \quad \qquad De-Chuan Zhan\\
	State Key Laboratory for Novel Software Technology, Nanjing University\\
	{\tt\small \{zhoudw, yehj\}@lamda.nju.edu.cn, zhandc@nju.edu.cn}
}

\maketitle
\thispagestyle{empty}

\footnotetext[2]{Correspondence to: Han-Jia Ye (yehj@lamda.nju.edu.cn)}

\begin{abstract}
	Traditional classifiers are deployed under closed-set setting, with both training and test classes belong to the same set. However, real-world applications probably face the input of unknown categories, and the model will recognize them as known ones. Under such circumstances, open-set recognition is proposed to maintain classification performance on known classes and reject unknowns. 
	The closed-set models make overconfident predictions over familiar known class instances, so that calibration and thresholding across categories become essential issues when extending to an open-set environment.
	To this end, we proposed to learn PlaceholdeRs for Open-SEt Recognition (\mame), which prepares for the unknown classes by allocating placeholders for both data and classifier.
	In detail, learning data placeholders tries to anticipate open-set class data, thus transforms closed-set training into open-set training. 
	Besides, to learn the invariant information between target and non-target classes, we reserve classifier placeholders as the class-specific boundary between known and unknown.
	The proposed \name efficiently generates novel class by manifold mixup, and adaptively sets the value of reserved open-set classifier during training.
	Experiments on various datasets validate the effectiveness of our proposed method.
\end{abstract}

\section{Introduction}

Recent years have witnessed the rapid development of supervised learning, aiming to obtain the knowledge of finite known classes. During the testing process, the well-trained model matches an incoming instance to the class with the highest posterior probability. However, this closed-world assumption comes to an end when the test set includes unseen categories~\cite{li2005open,bendale2015towards,zhou2016learnware,ye2018rectify}. Since it is impossible to cover all classes in the world as training set~\cite{mccarthy1981some,yang2019adaptive}, the model would treat all novel category instances as known ones.
As a result, the performance decays, which is unbearable in real-world applications. Open-set recognition~\cite{Scheirer2013TPAMI,bendale2016towards,yu2017open,perera2020generative} is thus proposed to conduct classification on known instances while at the same time detect those from unknown classes.

\begin{figure}[t]
	\begin{center}
		\includegraphics[width=1\columnwidth]{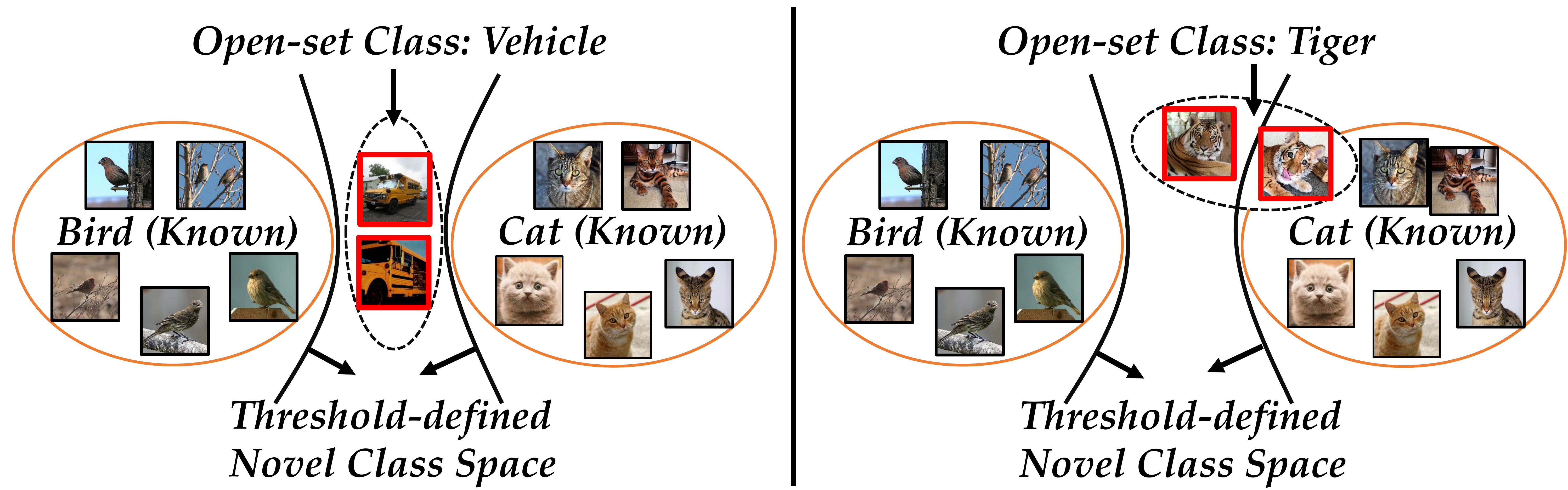}
	\end{center}
	\vspace{-2mm}
	\caption{ \small The drawback of threshold-based open-set recognition. It defines novel class space if confidence below a fixed threshold. Bird and cat are known classes, while vehicle and tiger are open-set classes in the left and right figures, respectively. 
		Since the distributions of the vehicle and tiger categories differ, it is hard to rely on a single threshold to recognize unknown classes with diverse characteristics. 
		The same threshold which well separates vehicles apart from known classes is not suitable for  tigers.
	} \label{figure:intro}
	\vspace{-3mm}
\end{figure}

Facing the unknown input of novel categories,
an intuitive way to separate known and unknown instances is to exert a threshold over the output probability~\cite{hendrycks2016baseline}. It assumes the model produces a higher probability for known classes than unknowns. However, deep learning methods tend to overfit the training instances and produce overconfident predictions~\cite{Scheirer2013TPAMI,guo2017calibration,hein2019relu}. As a result, the model would output a high probability even for an unknown class instance, making the threshold hard to tune. Besides, the class-compositions are diverse, as shown in Figure~\ref{figure:intro}. Since the semantic information of known classes differs in different tasks, it is hard to acquire an optimal threshold that suits all open-set tasks.  Consequently, it is urgent to calibrate closed-set classifiers.
Other methods try to foresee the distributions of novel classes and calibrate the output with open-set probability.
\firstcite{ge2017generative} proposed G-OpenMax, which utilizes GAN to generate unknown samples for training novel classifier. \firstcite{neal2018open} tried to generate images lying between decision boundaries as counterfactual instances. 
\firstcite{perera2020generative} combined self-supervision and augment input with generated open-set samples, which yields high disparity. 
These methods try to anticipate novel class distributions with generative models, and transform the closed-set training into open-set training.

The aim to boost open-set recognition can  be summarized as a calibration problem~\cite{guo2017calibration,ye2019learning}.
Firstly, to make the closed-set model prepare for unknown classes, \emph{data placeholders} of the novel class should be augmented and transform
open-set into closed-set. 
Secondly, to better separate known and unknown instances, overconfident predictions should be calibrated by reserving \emph{classifier placeholders} for novel classes.

Motivated by the problems above, we proposed to learn PlaceholdeRs for Open-SEt Recognition (\mame), aiming to calibrate open-set classifiers from two aspects. In detail, we augment the closed-set classifier with an extra \emph{classifier placeholder}, which stands for the class-specific threshold between known and unknown. We reserve the placeholder for open-set classes to acquire the invariant information between target and non-target classes.
Besides, to efficiently anticipate the distribution of novel classes, we consider generating \emph{data placeholders}, which mimic open-set categories with a limited complexity cost. Consequently, we can transform closed-set classifiers into open-set ones, and \emph{adaptively} predicts the class-specific threshold during testing. Experiments on various datasets validate the effectiveness of our proposed method on unknown detection and open-set recognition problems. Additionally, the visualization on decision boundaries indicates \name learns adaptive threshold for different class combinations.

In the following sections, we start with a brief review of related work, and then give the proposed \name and experiment results. After that, we conclude the paper.

\section{Related Work}
\bfname{Open-set Recognition.} There are  two  lines of work for open-set recognition, \ie, discriminative models and generative models~\cite{geng2020recent}.  Discriminative models can be further divided into traditional machine learning-based methods and deep learning-based methods. There has been much progress in traditional methods. Based on SVM, \firstcite{Scheirer2013TPAMI} proposed 1-vs-Set machine, aiming to create a slab in the feature space.  \firstcite{bendale2015towards} extended the nearest class mean classifier~\cite{mensink2013distance}, which calculates the distance between unknown and known class centers. \firstcite{zhang2016sparse} considered open-set recognition as a sparse representation learning problem, and tried to match the reconstruction error distributions with extreme value theory (EVT)~\cite{kotz2000extreme}. In recent years, deep learning-based methods have attracted more attention due to the powerful representation ability. \firstcite{bendale2016towards} first proposed to replace the softmax layer in the network with OpenMax, which calibrates the output probability with Weibull distribution. A similar work~\cite{shu2017doc} replaced the softmax layer with one-vs-rest units. \firstcite{yoshihashi2019classification} utilized the latent representation for reconstruction, enabling robust unknown detection with  known-class classification. These methods require a threshold to separate known and unknown, which face the challenge of threshold tuning.

Other works adopt generative models to anticipate the distribution of novel classes. G-OpenMax~\firstcite{ge2017generative} followed the guideline of OpenMax, which adopted a conditional generative network to synthesize unknown instances for network training. \firstcite{neal2018open} proposed counterfactual images for open-set recognition (OSRCI). OSRCI generates instances lying in the decision boundary, and augments the original dataset with these generated instances. Class conditional auto-encoder (C2AE)~\cite{oza2019c2ae} was proposed to tackle open-set recognition as a two-step problem, \ie, closed-set training and open-set training. C2AE used class conditioned auto-encoders with novel training and testing methodology. Recently, \firstcite{perera2020generative} utilized self-supervision and augmented the input image to learn richer features to improve separation between classes. These methods work with extra generative models, and transform the closed-set classifier into open-set classifier with the generated novel instances.

\bfname{Out-Of-Distribution Detection.} The problem of out-of-distribution detection~\cite{hendrycks2016baseline,liang2017enhancing,bai2020decaug}, anomaly detection~\cite{chandola2009anomaly,liu2008isolation,pang2020deep} and novelty detection~\cite{hodge2004survey,wei2019multiple,zhou2021detecting,mu2017classification,zhu2018multi} are related topics to open-set recognition. They can be viewed as the unseen class detector in the open-set recognition problem. The essential difference between them and open-set recognition lies in that they are a binary classification problem. Since out-of-distribution detection methods are not designed for classifying  known classes, they are not proper for open-set recognition.

\section{From Closed-set to Open-set Recognition}
In this section, we introduce the definition of closed-set classification and open-set recognition. Besides,  we show how to train a closed-set classifier, and the limitations when extending it into open-set recognition.
\subsection{Closed-Set Classification}
Traditional closed-set classifiers are trained with $\mathcal{D}_{tr}= \left\{ \left( \x _ { i }  , y _ { i }  \right) \right\} _ { i = 1 } ^ { L }$ and tested with $\mathcal{D}_{te}= \left\{ \left( \x _ { i }  , y _ { i }  \right) \right\} _ { i = 1 } ^ { M }$, where $ { \x_i } \in \mathbb{R} ^ { D }$ is a training instance, and  $y_i  \in Y  =\{ 1,2 , \ldots , K  \}$ is the associated class label. In the closed-set assumption, $\mathcal{D}_{tr}$ and $\mathcal{D}_{te}$ are drawn from the same distribution $\mathcal{D}$. An algorithm should fit a model $f(x): X \rightarrow Y$, which minimizes the expected risk:
\begin{equation}
	f^{*}=\argmin _{f \in \mathcal{H}} \mathbb{E}_{(\x, y) \sim \mathcal{D}_{te}} \mathbb{I}(y\neq f(\x)) \,,
	\label{eq:riskclose}
\end{equation}
where $\mathcal{H}$ is the hypothesis space, $\mathbb{I}(\cdot)$ is the indicator function which outputs $1$ if the expression holds and $0$ otherwise.
Assume the model $f$ is composed of embedding function $\phi(\cdot):\mathbb{R}^{D} \rightarrow \mathbb{R}^{d}$ and linear classifier $W\in\mathbb{R}^{d\times K}$, \ie, $f(\x)=W^{\top}\phi(\x)$. We denote the $k$-th column of $W$ as $\bm{w}_k\in\mathbb{R}^{d}$, \ie,  $W=[\bm{w}_1,\ldots,\bm{w}_K]$, thus the output logits of  class $k$ is $\bm{w}_k^{\top}\phi(\x)$. Generally, it can be optimized with cross-entropy to gain the discrimination among known classes. A validation set $\mathcal{D}_{val}$ drawn from closed set distribution $\D$ can be used to measure the closed-set performance.

\subsection{Open-Set Recognition}
Facing the emergence of open-set classes, the model is still trained with $\mathcal{D}_{tr}= \left\{ \left( \x _ { i }  , y _ { i }  \right) \right\} _ { i = 1 } ^ { L }$. However, now $\hat{\mathcal{D}}_{te}$ is filled with instances of novel categories, \ie,   $\hat{\mathcal{D}}_{te}= \left\{ \left( \x _ { i }  , y _ { i }  \right) \right\} _ { i = 1 } ^ { N }$, where  $y_i \in \hat{Y}  =\{ 1, \ldots , K , K + 1\}$ is the associated class label. Note that class $K+1$ is a group of novel categories, which may contain more than one class.
Since there is no side-information in the training set, and we are unable to decompose the novel class group into sub-categories. An optimal open-set classifier minimizes the expected risk~\cite{yu2017open}:
\begin{equation}
	\hat{f}^{*}=\argmin _{f \in \mathcal{H}} \mathbb{E}_{(\x, y) \sim \hat{\mathcal{D}}_{te}} \mathbb{I}(y\neq f(\x)) \,,
	\label{eq:riskopen}
\end{equation}
Since $\hat{\mathcal{D}}_{te}$ is composed of known classes and open-set class. The overall risk aims to classify known classes and meantime detect the unknown categories as class $K+1$. Traditional classifiers predict the instance with highest posterior probability, \ie, $\hat{y}=\argmax_{k=1,\cdots,K}\w_k^{\top}\phi(\x)$. However, since the model has not seen instances from open-set, it always predicts the lowest probability on class $K+1$. As a result, directly employing closed-set classifiers into open-set recognition will predict all novel instances into known categories, yielding unsatisfactory performance in open-set recognition.

Consequently, an intuitive way to adopt closed-set classifier for open-set recognition is thresholding~\cite{hendrycks2016baseline}. Taking the max output probability as confidence score, \ie, $conf=\max_{k=1,\ldots,K}\w_k^{\top}\phi(\x)$.  It assumes the model is more confident of closed-set instances than open-set. We can extend closed-set classifier by:
\begin{equation}	
	\begin{aligned}
		\hat{y}=\left\{
		\begin{array}{ll}
			\argmax\limits_{k=1,\cdots,K}\w_k^{\top}\phi(\x) & conf> th \\
			K+1 & \text{otherwise}
		\end{array}\right.
		\,,\label{eq:threshold}
	\end{aligned}
\end{equation}
where $th$ is the threshold. However, due to the overconfidence phenomena of deep neural networks, the output confidence of known and unknown is both close to~$1$~\cite{bendale2016towards}. As a result, tuning a threshold that well separates known from unknown is hard and time-consuming. Furthermore, since the relationship between known and unknown may be different, the threshold in Eq.~\ref{eq:threshold} relies on the essential similarity between known and unknown, which differs in diverse known-unknown compositions. To conclude,  the closed-set classifier should be equipped with an extra calibration process to suit open-set recognition requirements.

\section{  Learning Placeholders for Open-Set Recognition}
Facing the difficulty of closed-set classifier calibration, we need  placeholders to prepare the closed-set model for novel classes. The key idea of \name is to design placeholders in two aspects, \ie, data placeholders that anticipate novel classes, and classifier placeholders that separate known from unknown. Learning data placeholders aims to  mimic the emergence of novel classes, and transform closed-set training into open-set training. Reserving classifier placeholders for novel classes seeks to augment the closed-set classifier with dummy classifier, which adaptively outputs the class-specific threshold to separate known and unknown.
\begin{figure*}[t]
	\begin{center}
		\subfigure[Learning classifier placeholders]
		{\includegraphics[width=1.05\columnwidth]{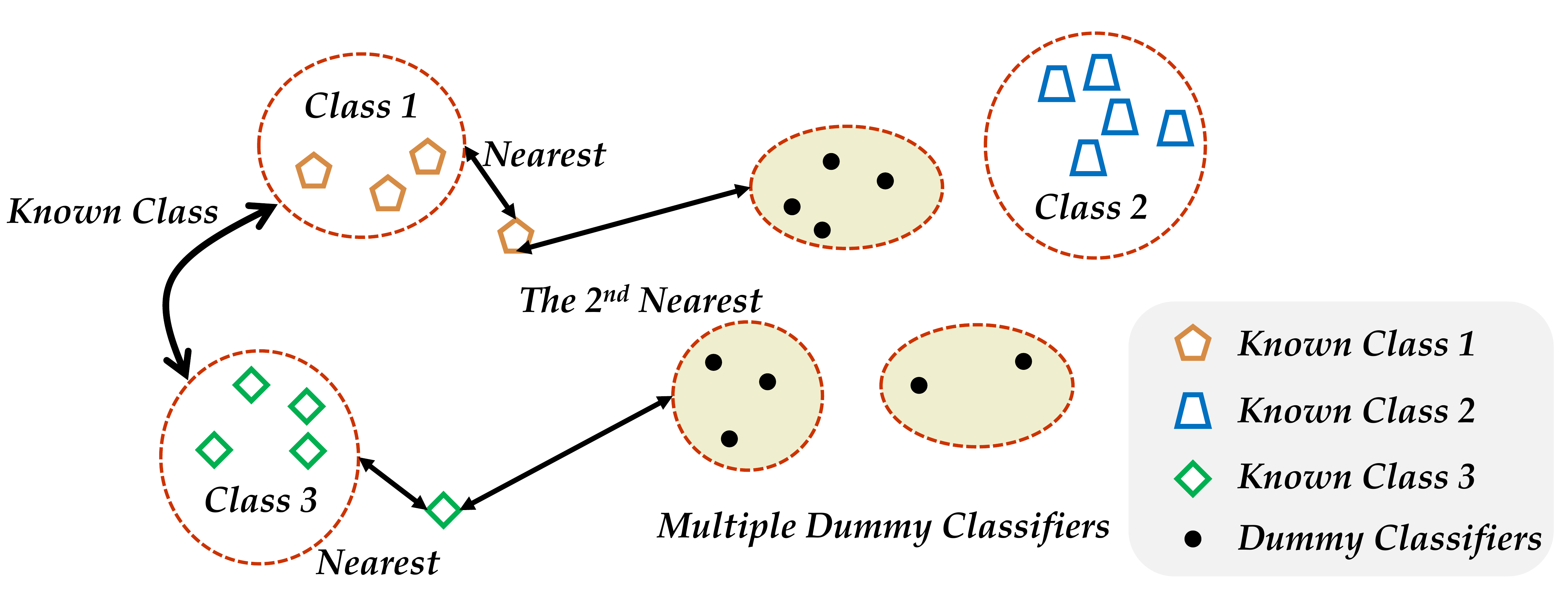}\label{figure:2a}}
		\subfigure[Learning data placeholders]
		{\includegraphics[width=1.\columnwidth]{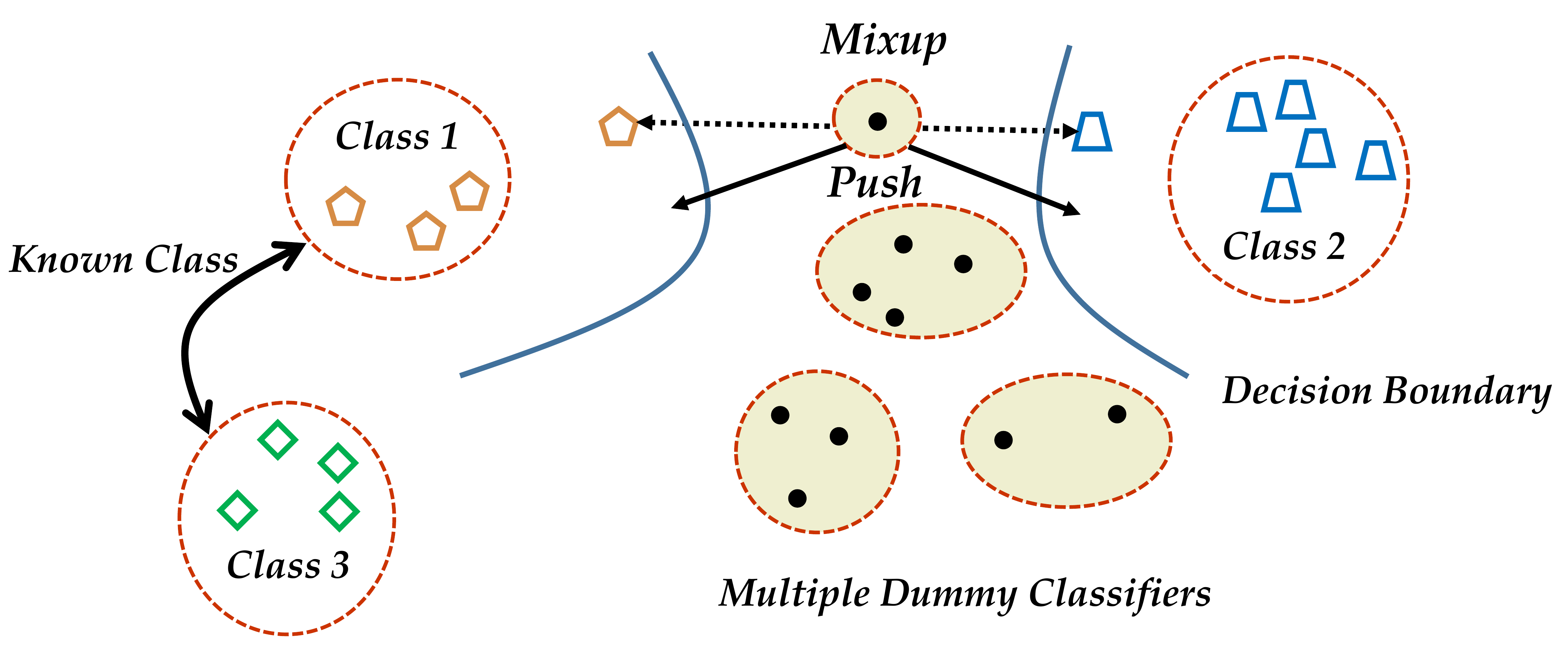} \label{figure:2b}}
	\end{center}
	\vspace{-3mm}
	\caption{Illustration of proposed \mame. The left figure corresponds to optimizing the  classifier placeholders by Eq.~\ref{eq:dummyloss} to output the second-largest probability, thus  place them between target and non-target classes. The right figure corresponds to anticipating novel class patterns by manifold mixup as Eq.~\ref{eq:lmix}, which generates data placeholders near the manifold of the decision boundary, and pushing the decision boundary much tighter. } \label{figure:method}
	\vspace{-3mm}
\end{figure*}
\subsection{Learning Classifier Placeholders}
To handle the diverse compositions of known-unknown categories, we need to extract invariant information from the target and non-target classes. 
Reserving classifier placeholders aims to arrange an extra dummy classifier and optimize it to represent the threshold between known and unknown.  
Assume we have the well-trained closed-set classifier; we first augment the output layer with an extra dummy classifier:
\begin{equation}
	\begin{aligned}
		\hat{f}(\x)=[W^{\top}\phi(\x),\hat{\w}^{\top}\phi(\x)] \,.
	\end{aligned}\label{eq:augment}
\end{equation}
Note that the dummy classifier $\hat{\w}$ shares the same embedding $\phi(\cdot)$ with the closed-set classifier, and only create an extra linear layer $\hat{\w}\in\mathbb{R}^{d\times 1}$. 
These augmented logits will then pass the softmax layer to produce the posterior probability.
The definition of dummy classifier is to well separate known and unknown, which is a dynamic threshold depend on input~$\x$, acting as the classifier placeholder. 
To this aim, we fine-tune the model and make the dummy classifier to output the second-largest probability for known instances. Through this way, the invariant information between known class classifier and dummy classifier can be transferred into the detection process. 
Since the current output is augmented with dummy classifier, the classification loss can be expressed as:
\begin{equation}
	l_{1}=\sum_{(\x,y)\in \mathcal{D}_{tr}}\ell(\hat{f}(\x),y)+\beta \ell(\hat{f}(\x)\setminus y,K+1) \,,
	\label{eq:dummyloss}
\end{equation}
where $\ell$ can be cross-entropy or other loss. The first item corresponds to optimizing the augmented output to match the ground truth, and maintaining performance in the closed-set. In the second term, $\hat{f}(\x)\setminus y$ means removing the probability of ground truth label, \ie, set the predicted probability of ground-truth ${\w}_y^{\top}\phi(\x)$ to $0$. The second item matches the masked-probability with class $K+1$, forcing the dummy classifier to output the second-largest probability. 
With the help of Eq.~\ref{eq:dummyloss}, the model learns to both correctly classify known instances and train the dummy classifier to place between the target and non-target classes. Note that loss is calculated with all training data $\mathcal{D}_{tr}$, which does not need novel class instances.

\noindent\bfname{Effects of dummy classifiers: }The explanation of Eq.~\ref{eq:dummyloss} is shown in Figure~\ref{figure:2a}. Eq.~\ref{eq:dummyloss} creates a way to calibrate the closed-set classifier, where the first item aims to push the instance towards its corresponding cluster to maintain correct classification. 
The second term seeks to relate the instance with the dummy classifier in the center space, and control the distance to the dummy classifier to be the second nearest among all class centers. As a result, it seeks a trade-off between correctly classifying closed-set instances and reserving novel classes' probability as the classifier placeholder. In the training process, it learns to place between target class and non-target classes. When faced with novel classes, the prediction of dummy classifier would be high since all known classes are non-target ones. As a result, it acts as the instance-dependent threshold which can well fit every known class.

\noindent\bfname{Learning multiple dummy classifiers:} As discussed in Eq.~\ref{eq:riskopen}, the $K+1$ class  may be composed of more than one novel class. We can  learn more dummy classifiers by augmenting  $\hat{W}\in\mathbb{R}^{d\times C}$, where $C$ corresponds to the number of dummy classifiers. Under such circumstance, Eq.~\ref{eq:augment} turns into augment $f(x)$ with the highest dummy logit, \ie,  $\hat{f}(\x)=[W^{\top}\phi(\x),\max\limits_{k=1,\cdots,C}\hat{\w}_{k}^{\top}\phi(\x)]$, which only considers the nearest dummy classifier.  With the help of multiple dummy classifiers, the model would adaptively choose the nearest dummy classifier to optimize. Learning multiple classifier placeholders boosts the variety of dummy classifiers from a united group into several scattered clusters. As a result, it leads to smoother decision boundaries, which in turn facilitates open-set recognition.

\subsection{Learning Data Placeholders }
The target of learning data placeholders is to change closed-set training into open-set training. The synthesized data placeholders should have two main characters, \ie, the distribution of these instances seems novel, and the generating process should be quick. Traditional generative-based open-set models tend to utilize powerful generative models to mimic novel patterns~\cite{neal2018open,perera2020generative}. 
However,  the distribution of natural images is hard  to modeling~\cite{yoshihashi2019classification}. 
To this end, we provide a simple yet effective way to anticipate novel class instances without any extra time complexity. 

Taking the above two characters into consideration, we mimic novel patterns with manifold mixup~\cite{verma2019manifold}. Assume the embedding module $\phi(\cdot)$ of the model can be decomposed by the middle hidden layer: $\phi(\x)=\phi_{post}(\phi_{pre}(\x))$, where $\phi_{pre}$ corresponds to the pre-layers before middle layer, which maps input into the hidden representation. Correspondingly, $\phi_{post}$ maps the hidden representation into the final embedding $\phi(\x)$.  We choose two instances from different class, and mix them up at the middle layer:
\begin{equation}\label{eq:mixup}
	\tilde \x_{pre}=\lambda \phi_{pre}(\x_{i})+(1-\lambda) \phi_{pre}(\x_{j}),  y_i \neq y_j \,,
\end{equation} 
where $\lambda \in [0,1]$ is sampled from Beta distribution. The mixed $\tilde x_{pre}$ will then pass the later layers, yielding $\phi_{post}(\tilde x_{pre})$. Considering the interpolation between two different clusters are often regions of low-confidence predictions~\cite{verma2019manifold}, \ie, places of non-target classes. As a result, we can treat the embedding $\phi_{post}(\tilde x_{pre})$ as the embedding of open-set classes, and train them as novel ones:
\begin{equation}
	\begin{aligned}
		l_{2}=\sum_{(\x_i,\x_j)\in \mathcal{D}_{tr}}\ell([W,\hat{\w}]^{\top}\phi_{post}(\tilde \x_{pre}),K+1) \\
		\tilde x_{pre}=\lambda \phi_{pre}(\x_{i})+(1-\lambda) \phi_{pre}(\x_{j}), y_i \neq y_j \,.
	\end{aligned}\label{eq:lmix}
\end{equation}
Note that we do not combine all possible pairs of $\left(\x_{i},\x_{j}\right)$ in the whole dataset to form $\tilde x_{pre}$. On the contrary, these combinations are produced within mini-batches~\cite{zhang2018mixup,verma2019manifold}, \ie, once we get the training batch of size $B$, another order of instances can be derived by shuffling this mini-batch. We then mask the pairs of the same class, and conduct manifold mixup with the pairs from different classes. As a result, the calculation complexity is of the same magnitude as vanilla training and would not cost extra time.

\noindent\bfname{Effects of data placeholders:} It is obvious that Eq.~\ref{eq:lmix} does not consume extra time complexity, which generates novel instances lying between the manifold of decision boundaries. Besides, manifold mixup can better generate novel patterns in the improved embedding space leveraging interpolations of deeper hidden representations, which better stands for the novel distribution. The visualization of data placeholders is shown in Figure~\ref{figure:2b}, where the mixed instances push the decision boundary in the embedding space towards the composition class $y_{ i }$ and $y_j$. With the help of data placeholders, the embeddings of known classes would be much tighter, leaving more place for novel classes.

\noindent\bfname{Discussion about vanilla mixup:}
Vanilla mixup~\cite{zhang2018mixup,tokozume2018between} aims to produce augmented instances with linear interpolations between two known ones in the input space: $\tilde{\x}=\lambda \x_{i}+(1-\lambda) \x_{j} $. However, a fatal problem may occur that mixed $\tilde{\x}$ in the input space may situate near another class $y_k$ between $y_i$ and $y_j$, \ie, $\lambda \x_{i}+(1-\lambda) \x_{j} \approx \x_k$. Since we will treat mixed instances as class $K+1$, optimizing such $\tilde{\x}$ harms the discriminability among closed-set classes and semantic information of original inputs. On the contrary, \cite{verma2019manifold} proved that manifold mixup can move the decision boundary away from the data in all directions, resulting in a compact embedding space. Besides, since the features in the input space are fixed, the generated novel patterns by vanilla mixup are thus deterministic and cannot be optimized. By contrast, the generated novel patterns can be optimized with the embedding module $\phi(\cdot)$.

\subsection{Calibration and Guideline for Implementation }
Previously, we talk about the elements in \mame, and how they are separately trained. We then discuss the combination of these two parts and extra calibration tricks to further improve the classifier.

Rethinking the overconfidence problem of closed-set classifier, we should make dummy classifier output the same magnitude logits as closed-set classifier.
To fully explore the magnitude relationship between closed-set classifier  $W^{\top}\phi(\x)$ and dummy classifier $\hat{\w}^{\top}\phi(\x)$, we consider a further calibration step. With the help of validation set $\mathcal{D}_{val}$, we can calculate the highest logit of known classes  and  dummy classifier. To calibrate them into same magnitude, we calculate the difference of them:
$\max\limits_{k=1,\cdots,K} \w_k^{\top}\phi(\x)-\max\limits_{k=1,\cdots,C} \hat{\w}_k^{\top}\phi(\x) \,,
$
and divide it into several equal intervals as $bias$. The calibrated logits is $
[W^{\top}\phi(\x),\hat{\w}^{\top}\phi(\x)+bias],
$
we obtain the best $bias$ by ensuring $95\%$ of validation data in $\mathcal{D}_{val}$ to be recognized as known~\cite{sun2020conditional,perera2020generative}. 
As a result, the dummy logit will be further calibrated to the same magnitude as closed-set classifiers with bias tuning. Note that the bias tuning is conducted with the validation set $\mathcal{D}_{val}$ drawn from the same distribution as $\mathcal{D}_{tr}$ and with no access to the unknown dataset.

In the training process of \name, we first augment the dummy classifiers with pretrained closed-set classifiers. After that, we separate each batch into two equal parts, and separately calculate $l_1$ and $l_2$ over corresponding parts. The manifold mixup loss $l_2$ is calculated by shuffling the mini-batch and mixup instances of different classes, which takes no extra time.
After the training process, we utilize the validation set to get the calibration bias. The guideline for implementation is shown in Algorithm~\ref{alg1}. The optimal bias would be used to accumulate dummy classifier logits during open-set testing.
\begin{algorithm}[t]
	\small
	\caption{ Training \name for open-set recognition }
	\label{alg1}
	{\bf Input}:   
	Closed-set classifier: $f$; Dummy classifier number: $C$; \\
	Closed-set training set: $\mathcal{D}_{tr}= \left\{ \left( \x _ { i }  , y _ { i }  \right) \right\} _ { i = 1 } ^ { L }$;

	{\bf Output}: 
	Open-set classifier: $\hat{f}$; 
	Calibration bias: $bias$;

	\begin{algorithmic}[1]{
			\STATE Initialize $\hat{W}\in\mathbb{R}^{d\times C}$ as dummy classifier;
			\STATE Augment the output of  $f$ with dummy classifier as Eq.~\ref{eq:augment};
			\REPEAT
			\STATE Get a batch of training instance $ \left\{ \left( \x _ { i }  , y _ { i }  \right) \right\} _ { i = 1 } ^ { B }$;
			\STATE Separate the batch into two parts with equal size; 
			\STATE Calculate the dummy loss on the first part $l_{1}\leftarrow$ Eq.~\ref{eq:dummyloss};
			\STATE Conduct manifold mixup on the second part, calculate mix loss $l_{2}\leftarrow$ Eq.~\ref{eq:lmix};
			\STATE Calculate the overall loss $l_{total}=l_{1}+\gamma*l_{2}$;
			\STATE Obtain derivative and update the model;
			\UNTIL reaches predefined epoches}
		\STATE Calibrate the output of dummy classifier over validation set by ensuring $95\%$ of validation data recognized as known;
	\end{algorithmic}
	
\end{algorithm}

\begin{table*}[t] 
	\centering{
		\caption{Unknown detection performance in terms of the mean AUC. Results are averaged among five randomized trials. We report the full table with standard deviation in the supplementary.}
		{\begin{tabular}{l|c|c|c|c|c}
				\addlinespace
				\toprule
				{Methods} &{SVHN}  &{CIFAR10}  &{CIFAR+10}  &{CIFAR+50}    
				& Tiny-ImageNet \\
				\midrule
				Softmax &  88.6 & 67.7    & 81.6 & 80.5    &57.7  \\
				
				OpenMax~\cite{bendale2016towards} & 89.4 &  69.5 & 81.7 & 79.6    &57.6 \\
				
				G-OpenMax~\cite{ge2017generative} & 89.6 &  67.5 & 82.7 & 81.9   &58.0 \\
				OSRCI~\cite{neal2018open} & 91.0 &  69.9    & 83.8 & 82.7    &58.6 \\
				
				C2AE~\cite{oza2019c2ae}& 89.2 &  71.1    & 81.0 & 80.3   &58.1 \\
				
				%CROSR~\cite{yoshihashi2019classification} & 89.9 $\pm$ 1.8&  \nrr   & \nrr&  \nrr   &58.9 $\pm$ \nrr\\
				
				GFROSR~\cite{perera2020generative} & 93.5 &  83.1   & 91.5 & 91.3   &64.7 \\
				
				\midrule
				\name &\bf 94.3  &\bf 89.1 & \bf 96.0 & \bf95.3&  \bf69.3 \\
				\bottomrule
			\end{tabular}\label{table:unknown}
	}}
\end{table*}

\section{Experiment}
In this section, we compare \name on benchmark datasets with state-of-the-art methods. We separately evaluate the performance of unknown detection and open-set recognition tasks. Ablations show the model robustly tackles datasets with different complexity, and visualizations indicate \mame's ability in adaptive threshold choosing.

\subsection{Unknown Detection}
Following the protocol defined in~\cite{neal2018open,yoshihashi2019classification,perera2020generative}, we evaluate the performance of related methods. Several classes are sampled from a multi-class dataset and the others are viewed as open-set class~\cite{neal2018open}. We simulate the sampling process over five trials~\cite{neal2018open}, and report the mean results. The commonly used metric \ie, average area under the ROC curve (a.k.a AUC)  evaluates the recognition performance. Since real-world scenarios are complex, where ratio of seen and unseen differs in diverse tasks, we utilize openness\footnote{There are two similar definitions of openness, and we follow~\cite{neal2018open,perera2020generative}.}~\cite{perera2020generative,neal2018open,Scheirer2013TPAMI,geng2020recent} to represent the complexity of the open-set task:
\begin{equation}
	\text{Openness}=1-\sqrt{\frac{N_{train}}{N_{test}}} \,,
	\label{eq:openness}
\end{equation}
where $N_{train}\text{ and }N_{test}$ corresponds to the number of classes in training set and testing set. As we discussed in the preliminaries, $N_{train}=K$. We compare to other methods on the following benchmark datasets:

\begin{itemize}
	\item{ \bfname{SVHN}~\cite{netzer2011reading} and \bfname{CIFAR10}~\cite{krizhevsky2009learning}}: There are total 10 classes in these datasets. SVHN contains street view house numbers, and CIFAR10 contains images of vehicles and animals. We randomly sample six classes to be known and the other four classes to be open-set classes. The openness of these tasks is $22.54\%$.
	\item \bfname{CIFAR+10}~\cite{neal2018open} and \bfname{CIFAR+50}: To create a dataset with higher openness, four classes from CIFAR10 is sampled as known class. Besides, 10 and 50 classes are sampled from CIFAR100 as open-set classes, yielding CIFAR+10 and CIFAR+50. The openness for them are $46.55\%\text{ and }72.78\%$, respectively.
	\item{\bfname{Tiny-ImageNet}~\cite{le2015tiny}}: Tiny-ImageNet is a subset of ImageNet~\cite{deng2009imagenet} with 200 classes. We sample 20 classes as known and the other 180 as open-set classes. The openness is $68.37\%$ for this dataset.
\end{itemize}

\begin{table}[t] 
	\centering{
		\caption{Closed-set accuracy between the plain CNN (closed-set classifier) and \mame. Although \name aims at learning placeholders for novel classes, there is no significant degradation in closed-set accuracy.}
		{\begin{tabular}{l|c|c|c}
				\addlinespace
				\toprule
				{Methods} &{SVHN}  &CIFAR10 &  Tiny-ImageNet\\
				\midrule
				Plain CNN & 96.5  &  92.8 &52.2  \\	
				\midrule
				\name  & 96.4 & 92.6 & 52.1\\
				\bottomrule
			\end{tabular}\label{table:close}
		}
	}
\end{table}

\begin{table*}[t] 
	\centering{
		\caption{Open-set recognition results on CIFAR10 with various outliers added to the test set as unknowns. The performance is evaluated by macro F1 in 11 classes (10 known classes and unknown). 	}
		{\begin{tabular}{l|c|c|c|c}
				\addlinespace
				\toprule
				{Methods} &{ImageNet-crop}  &{ImageNet-resize}  &{LSUN-crop}  &{LSUN-resize}    
				\\
				\midrule
				Softmax & 63.9&  65.3   & 64.2& 64.7  \\
				%OpenMax (CVPR16)~\cite{bendale2016towards}
				OpenMax~\cite{bendale2016towards} & 66.0 &  68.4   & 65.7&66.8    \\
				OSRCI~\cite{neal2018open} & 63.6&  63.5 & 65.0 &64.8\\
				LadderNet+Softmax~\cite{yoshihashi2019classification} & 64.0&  64.6   & 64.4& 64.7\\
				LadderNet+OpenMax~\cite{yoshihashi2019classification} & 65.3 &  67.0   & 65.2 & 65.9   \\
				%(CVPR19)
				DHRNet+Softmax~\cite{yoshihashi2019classification}& 64.5&  64.9  & 65.0& 64.9   \\
				%(CVPR19)
				DHRNet+OpenMax~\cite{yoshihashi2019classification} & 65.5&  67.5   & 65.6& 66.4   \\
				CROSR~\cite{yoshihashi2019classification} & 72.1&  73.5   & 72.0& 74.9   \\
				% (CVPR20)
				GFROSR~\cite{perera2020generative} & 75.7&  79.2  & 75.1& 80.5    \\		
				\midrule
				\name &\bf84.9 &\bf82.4&\bf86.7&\bf85.6 \\
				\bottomrule
		\end{tabular}}\label{table:cifarood}
	}
\end{table*}
We compare to the SOTA methods, \ie, Softmax, OpenMax~\cite{bendale2016towards}, G-OpenMax~\cite{ge2017generative}, OSRCI~\cite{neal2018open}, C2AE~\cite{oza2019c2ae}, CROSR~\cite{yoshihashi2019classification}, GFROSR~\cite{perera2020generative}:
{ \bfname{Softmax}}: utilizes the highest softmax probability as the confidence score for detection;	\bfname{OpenMax}: replaces softmax layer with OpenMax and calibrates the confidence to predict novel class;
{\bfname{G-OpenMax}}: trains conditional GAN to generate open-set instances and adopts OpenMax for recognition;
{\bfname{OSRCI}}: generates instances lying near the decision boundary;
{\bfname{C2AE}}: uses class conditioned auto-encoders with novel training and testing methodology;
{\bfname{CROSR}}: utilizes the latent representation learning for reconstruction, which enables robust unknown detection;
{\bfname{GFROSR}}: adopts  self-supervision and augments the input image to learn richer features to improve separation between classes;

We adopt the same network backbone and dataset splits as GFROSR~\cite{perera2020generative}. Table~\ref{table:unknown} shows the performance of unknown detection. We report the baseline performance from~\cite{perera2020generative,yoshihashi2019classification,neal2018open}. From Table~\ref{table:unknown}, we can infer that the results on traditional digital number datasets SVHN are almost saturated, and little progress can be further achieved. However, our proposed \name improves the recognition ability on natural images by a substantial margin, \ie, in the experiments on CIFAR10, we push forward $6\%$ than SOTA method GFROSR. We also improve the performance over CIFAR+10, CIFAR+50, and Tiny-ImageNet by $4\%$. We also provide the closed set accuracy in Table~\ref{table:close}, which indicates that learning \name does not sacrifice the model's discriminative ability in the closed-set classification.

\begin{table}[t] 
	\centering{
		\caption{Open-set recognition on MNIST with various outliers added to the test set as unknowns. We report macro F1 in 11 classes (0–9 and unknown). The results are cited from~\cite{sun2020conditional}.}
		{\begin{tabular}{l|c|c|c}
				\addlinespace
				\toprule
				{Methods} &{Omniglot}  &{MNIST-noise} &  Noise \\
				\midrule
				Softmax & 59.5 &  64.1&82.9 \\
				OpenMax & 68.0 & 72.0 &82.6\\
				CROSR &79.3&  82.7 &82.6    \\
				\midrule
				\name & \bf 86.2 & \bf87.4& \bf 88.2 \\
				\bottomrule
			\end{tabular}\label{table:mnistood}}
	}
\end{table}

\begin{figure}[t]
	\begin{center}
		\includegraphics[width=.9\columnwidth]{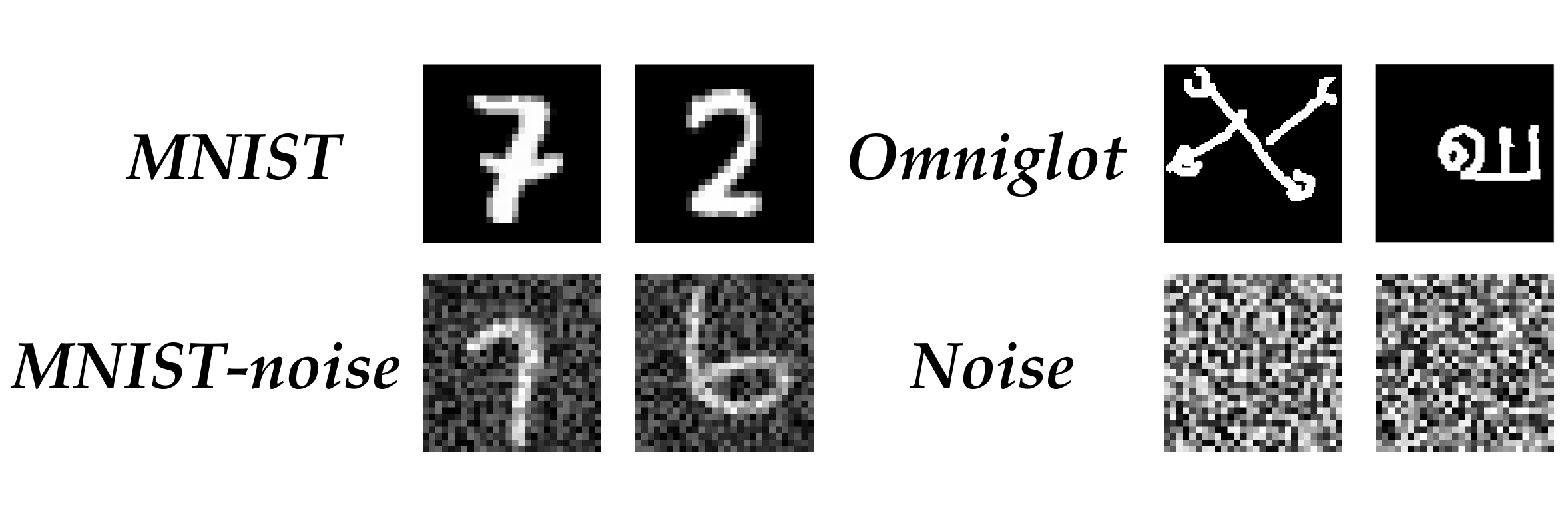}
	\end{center}
	\vspace{-3mm}
	\caption{Dataset example of original MNIST, Omniglot, MNIST-noise and Noise.} \label{figure:mnist}
	\vspace{-3mm}
\end{figure}

\subsection{Open-Set Recognition}
The ultimate goal of open-set recognition is to not only detect unseen classes, but also correctly classify known ones with superior performance. In this section, we validate the performance of our proposed \name with open-set recognition tasks. Following the protocol defined in~\cite{perera2020generative}, the models are trained by all training instances of the original dataset. While in the testing process, instances from another dataset are augmented to the original test set as open-set classes. In this setting, \emph{macro-averaged F1-scores} over all known classes and the augmented novel class is used to measure the performance. Two benchmark datasets, \ie, MNIST and CIFAR10 are used to simulate this setting.

We first consider MNIST, which is formed with digital numbers between $0-9$. According to~\cite{yoshihashi2019classification}, we choose the open-set classes from Omniglot~\cite{lake2015human}, MNIST-noise, and Noise. Omniglot is a dataset of alphabet characters, and Noise is randomly generated by sampling each pixel between $[0,1]$ from a uniform distribution. Superimposing MNIST images over Noise yields the MNIST-noise dataset, which is similar to the original MNIST. We show the dataset example in Figure~\ref{figure:mnist}. We make the known-unknown ratio 1:1 by setting the number of test examples from open-set to 10,000, the same as the number of MNIST test set. 
We evaluate the performance with macro F1 scores between ten known classes and one unknown class.
The results of open-set recognition are shown in Table~\ref{table:mnistood}.
It indicates that open-set recognition on Omniglot and MNIST-noise are more challenging among the three datasets, since they have higher openness than Noise. However, our proposed \name handles these scenarios with the best performance.

We also conduct experiments with CIFAR10, which is a dataset of vehicles and animals. We choose  open-set classes from ImageNet~\cite{deng2009imagenet} and LSUN~\cite{yu15lsun} according to~\cite{liang2017enhancing}. 
LSUN has a testing set of 10,000 images of 10 different scenes.
Since the image size of novel classes does not match CIFAR10, we design  ImageNet-crop, ImageNet-resize, LSUN-crop, and LSUN-resize to align input size. For `crop' datasets, we crop the original image into 32*32, and we resize the image into 32*32 for `resize' datasets. Like the experiments with MNIST, we set the size of novel instances to 10,000, yielding known-unknown ratio 1:1. 
We evaluate the performance with macro F1 scores between ten known classes and one unknown class.
The results concerning CIFAR10  are reported in Table~\ref{table:cifarood}. We can infer from Table~\ref{table:cifarood} that \name can handle open-set classes from diverse inputs and achieve better performance than SOTA methods.
\begin{figure}[t]
	\begin{center}
		\includegraphics[width=.88\columnwidth]{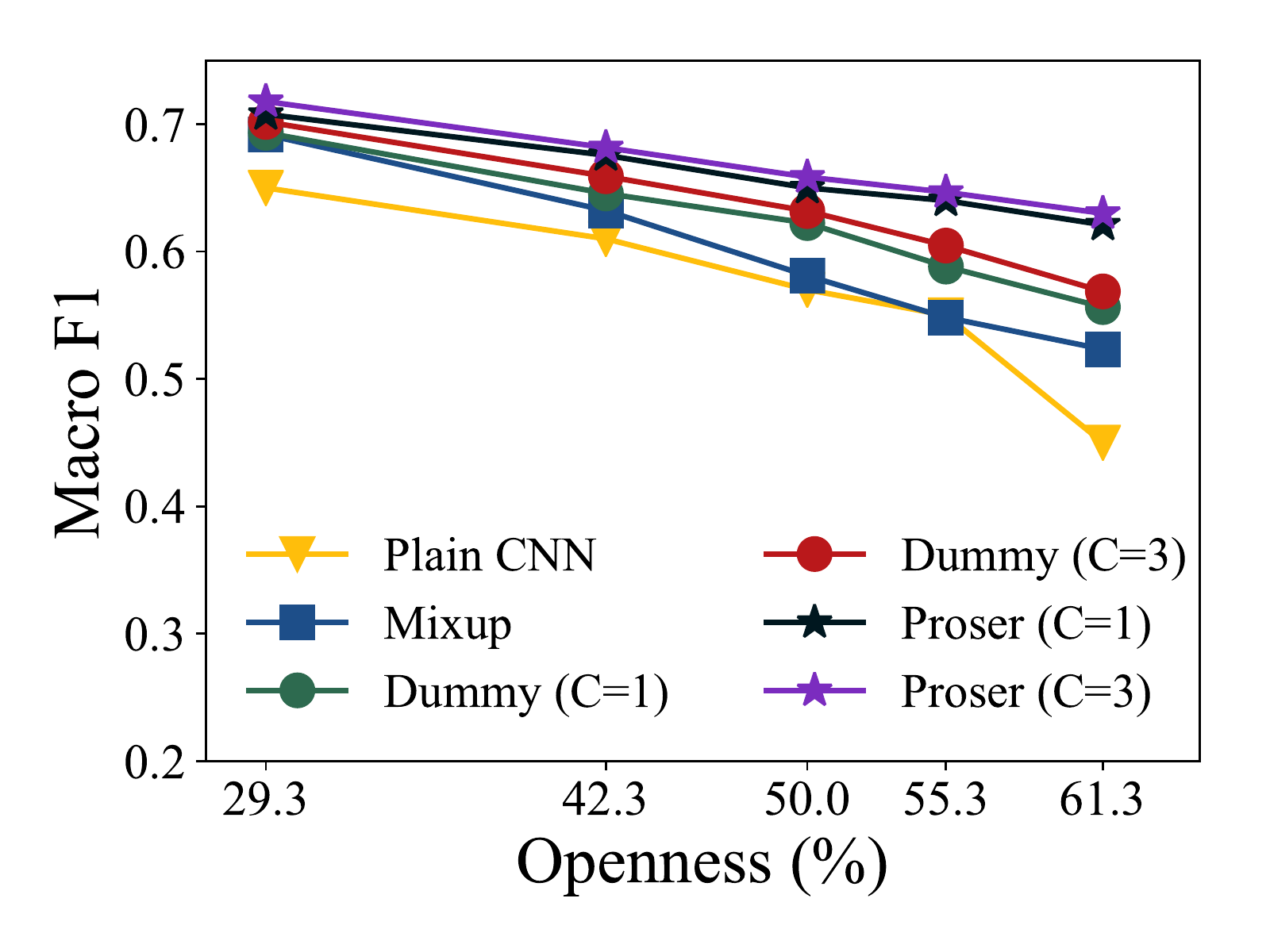}
	\end{center}
	\vspace{-5mm}
	\caption{Macro F1 against varying openness with different baselines for ablation analysis.}
	\vspace{-4mm} \label{figure:abalation}	
\end{figure}

\begin{figure*}[t]	
	\begin{center}
		\subfigure[\scriptsize  Thresholding when Class 4 is novel]
		{\includegraphics[width=.51\columnwidth]{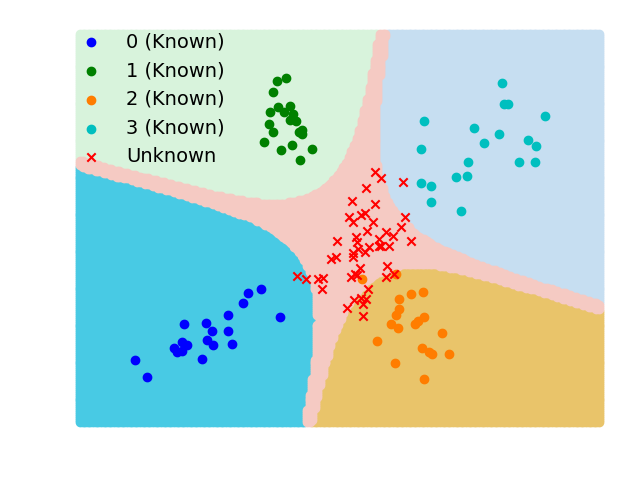}\label{figure:tsnea}}
		\subfigure[\scriptsize Thresholding when Class 5 is novel]
		{\includegraphics[width=.51\columnwidth]{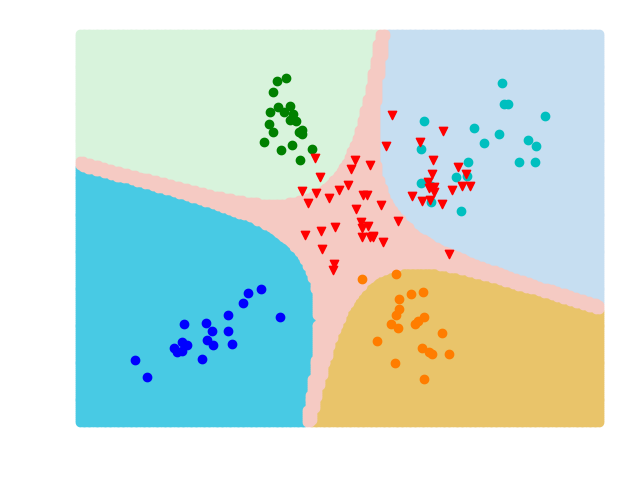}\label{figure:tsneb}}
		\subfigure[\scriptsize \name  when Class 4 is novel]
		{\includegraphics[width=.51\columnwidth]{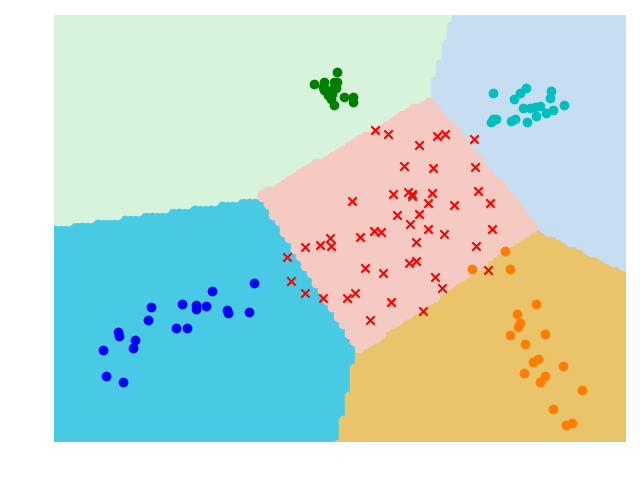}\label{figure:tsnec}}
		\subfigure[\scriptsize \name  when Class 5 is novel]
		{\includegraphics[width=.51\columnwidth]{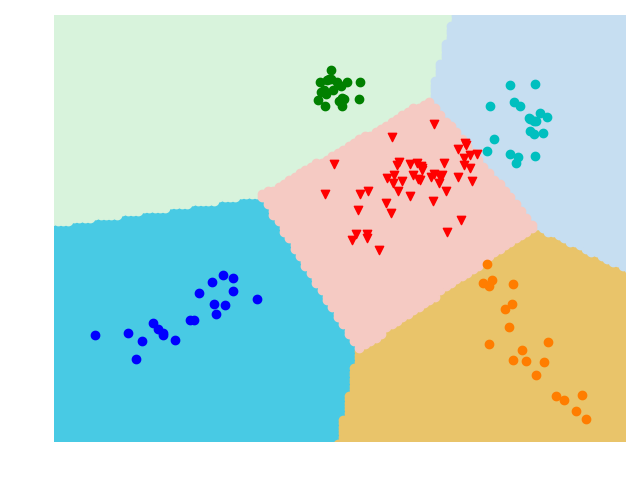}\label{figure:tsned}}
	\end{center}
	\vspace{-3mm}
	\caption{Visualization on open-set recognition of MNIST, known classes are visualized by dots with different colors,  and the unknown classes are denoted by red cross/triangle. The shadow region represents the classification boundary of the model. Thresholding methods define unknown space if confidence is below some threshold. As shown in (a) and (b), the unknown class (class 4 and class 5) may come from diverse distributions and hard to be detected with a uniform threshold.
		It indicates a suitable threshold to detect open-set class $4$ does not suit novelty from class $5$. While (c) and (d) show \name can bear the distribution change of novel classes with the help of adaptive threshold choosing. Best viewed in color. }\label{figure:tsne} 
	\vspace{-3mm}
\end{figure*}
\subsection{Ablation study}
In this section, we conduct an ablation study  and analyze each part's contribution with the CIFAR100 dataset.  CIFAR100 contains  100 classes, and we change the composition of known and unknown by varying openness in Eq.~\ref{eq:openness}. We randomly choose 15 out of 100 classes as known, and switch the number of unknown classes between $\{15, 30, 45, 60, 85\}$. The openness of these datasets is within the limits of $29.29\%$ and $61.28\%$. We evaluate the performance by macro F1-scores over 15 known classes and unknown. The results are shown in Figure~\ref{figure:abalation}.

In the figure, `Plain CNN' stands for thresholding  softmax probabilities as Eq.~\ref{eq:threshold}. `Mixup' stands for learning data placeholders only, and `Dummy' stands for learning classifier placeholders only. The parameter $C$ corresponds to the dummy classifier number.
The trend of all methods with openness increasing declines, since the task becomes more complex.   We can infer from the figure that adopting manifold mixup and dummy classifiers both improves the open-set recognition ability of plain CNN. The results indicate that data placeholders and classifier placeholders can both help calibration. Besides, learning more than one dummy classifier improves performance than one, indicating the diversity of dummy classifier matters. Furthermore, combining these two parts  can further improve the performance, which outperforms the others. Ablations validate that  placeholders help to calibrate the open-set classifier.

\subsection{Visualization of Decision Boundaries}

In this part, we visualize the learned decision boundaries on the MNIST dataset. Instances are shown in 2D by learning embedding module $\phi(\cdot):\mathbb{R}^{D} \rightarrow \mathbb{R}^{2}$, \ie, a linear layer attached to the CNNs as embedding.
We plot the comparison between traditional threshold-based methods and our \name in Figure~\ref{figure:tsne}. In each figure, four known classes (in dots) and one open-set class (in red cross/triangle) are visualized. A threshold-based method detects open-set class by Eq.~\ref{eq:threshold}, and we tune a suitable threshold for novel class 4 in Figure~\ref{figure:tsnea}. The same decision boundary (threshold) is adopted in \ref{figure:tsneb}, where known classes are the same while class 5 is the novel class. We can infer from  Figure~\ref{figure:tsnea} and~\ref{figure:tsneb} that the distribution of open-set classes differs, where a well-defined threshold for novel class 4 is not suitable for novel class 5. As a result, it is impossible to acquire an optimal threshold that well separates all kinds of unknown classes from known ones. 

For comparison, we show the decision boundaries of \name in Figure~\ref{figure:tsnec},~\ref{figure:tsned}. With the help of data placeholders and classifier placeholders, \name is able to acquire the invariant information between target and non-target classes. As a result, it learns to adaptively output the instance-specific threshold, and bears the distribution change of novel classes in the novel space. Figure~\ref{figure:tsne} validates the effectiveness of \name facing various kind of open-set classes.

\section{Conclusion}
In real-world applications, instances from unseen novel classes may be fed to closed-set classifiers and be miss-classified as known ones. Open-set recognition aims to simultaneously classify known classes and detect unknown ones. However, there are two main challenges in open-set recognition, \ie, how to anticipating novel patterns and how to compensate for the overconfidence phenomena. In this paper, we propose \name to calibrate the closed-set classifiers in two aspects. On the one hand, \name efficiently mimics the distribution of novel classes as data placeholders, and transforms closed-set training into open-set training. On the other hand, we augment the closed-set classifier with classifier placeholders, which adaptively separates the known form unknown, and stands for the class-specific threshold. The proposed \name efficiently generates novel class by manifold mixup, and adaptively sets the value of reserved open-set classifier. How to extend open-set recognition into stream data scenarios, and utilize the detected novel patterns are interesting future works.

\section*{Acknowledgments}

This research was supported by National Key
R\&D Program of China (2020AAA0109401), NSFC (61773198, 61921006,62006112), NSFC-NRF Joint Research Project under Grant 61861146001, Nanjing University Innovation Program for
Ph.D. candidate (CXYJ21-53), Collaborative Innovation Center of Novel Software Technology and Industrialization, NSF of Jiangsu Province (BK20200313).

{\small
\bibliographystyle{ieee_fullname}
\bibliography{scis}
}

%\pagebreak
\clearpage
\setcounter{section}{0}
\renewcommand{\thesection}{\Roman{section}}
\begin{center}
	\textbf{\large Supplementary Material }
\end{center}
%%%%%%%%%% Merge with supplemental materials %%%%%%%%%%
%%%%%%%%%% Prefix a "S" to all equations, figures, tables and reset the counter %%%%%%%%%%
\setcounter{equation}{0}
\setcounter{figure}{0}
\setcounter{table}{0}
\setcounter{page}{1}
\makeatletter

\section{Additional Experimental Results}
This section introduces some additional experiment results and then gives the implementation details.

\subsection{Sensitivity about Hyper-Parameters}

In this section, we conduct experiments to explore the influence of hyper-parameters with the CIFAR100 dataset. The implementation details are the same as the ablations of the main paper. We choose 15 classes out of 100 as known ones, and another 15 out of 85 are open-set categories, making known-unknown ratio $1:1$. The performance is measured with macro F1 over 15 known classes and unknown.

\begin{figure}[t]
	\begin{center}
		\subfigure[Trade-off parameter]
		{\includegraphics[width=.495\columnwidth]{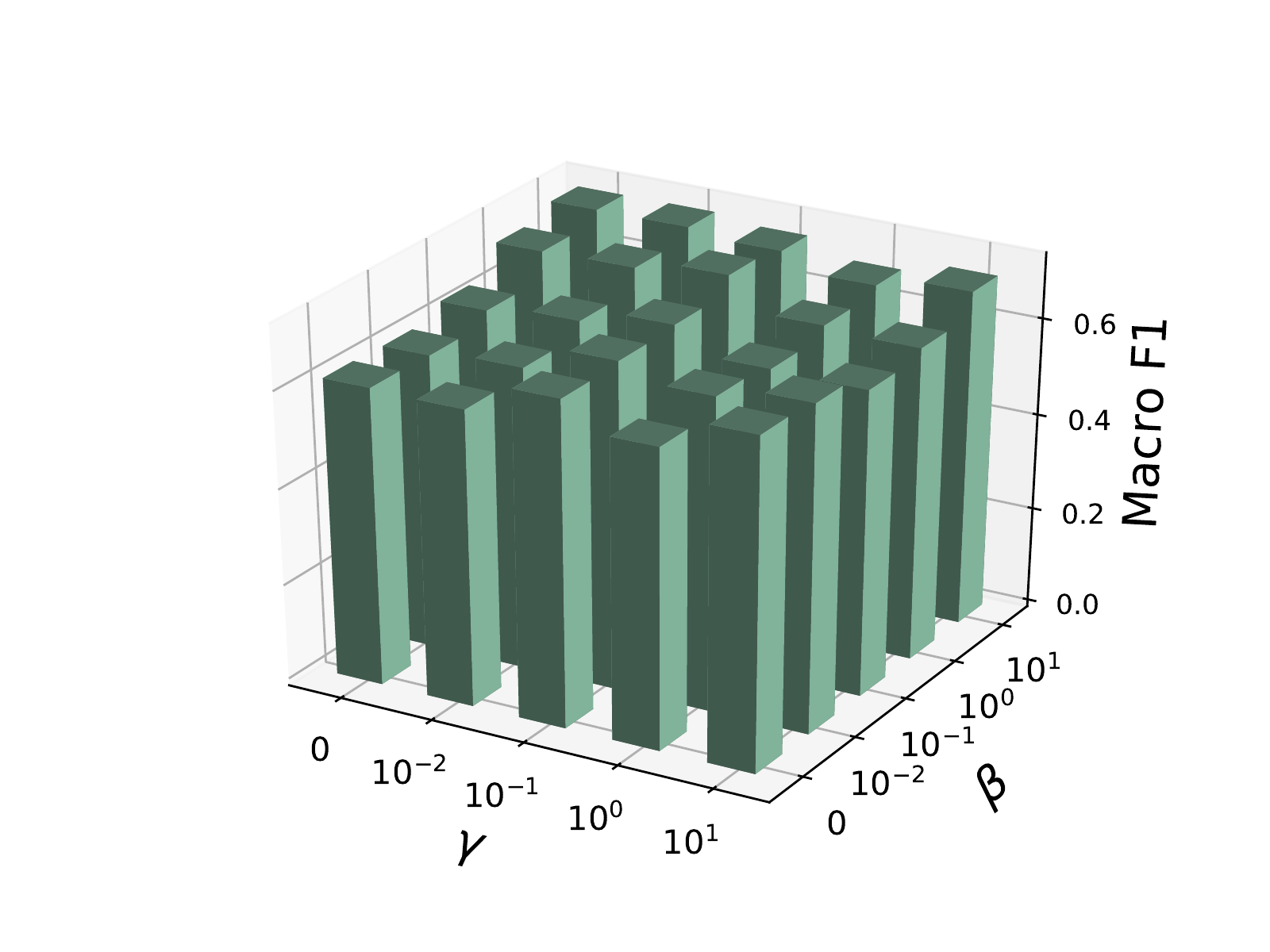}\label{fig:supp1a}}
		\subfigure[Dummy classifier number]
		{\includegraphics[width=.495\columnwidth]{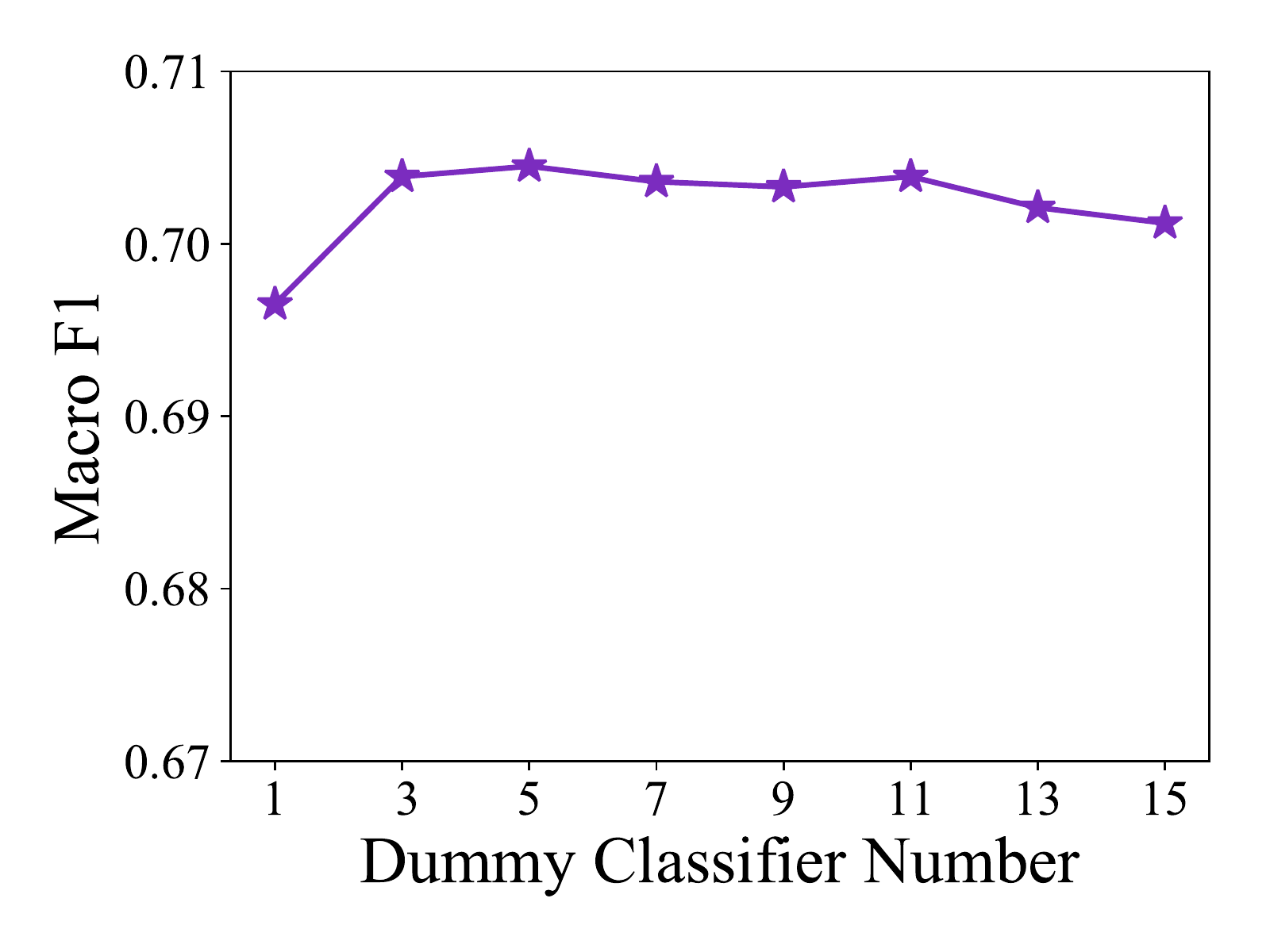}\label{fig:supp1b}}
		
	\end{center}
	\caption{Sensitivity of hyper-parameters on CIFAR100 dataset. We report macro F1 between 15 known classes and unknown.} \label{fig:suppparam}
\end{figure}

In the main paper, the overall loss is described as:
\begin{equation}
	l_{total}=l_{1}+\gamma*l_{2}\,,
\end{equation}
where $
l_{1}=\sum_{(\x,y)\in \mathcal{D}_{tr}}\ell(\hat{f}(\x),y)+\beta \ell(\hat{f}(\x)\setminus y,K+1) 
$ is the classifier placeholder loss, and $
l_{2}=\sum_{(\x_i,\x_j)\in \mathcal{D}_{tr}}\ell([W,\hat{\w}]^{\top}\phi_{post}(\tilde \x_{pre}),K+1) 
$ is the data placeholder loss, yielding three different parts. In this section, we report the effects of these hyper-parameters $\beta$ and $\gamma$ as well as dummy classifier number $C$ in Figure~\ref{fig:suppparam}. Considering the trade-off parameters are adopted to balance the loss of each parts, we tune them in range of $\{0,10^{-2},10^{-1},10^{0},10^{1}\}$. As a result, we can get 25 results, corresponding to each combination from the set.

Note that we have provided an ablation study in the main paper, where `Mixup' stands for only equipping the plain CNN with data placeholders, \ie, $\gamma\rightarrow\infty$. Correspondingly, `Dummy' stands for only training dummy classifiers, \ie, $\gamma=0$. The results in Figure~\ref{fig:supp1a} are consistent with the former conclusions that only employ part of the placeholder, \ie, data or classifier, is not enough to produce the best performance. Additionally, we can observe that $\beta=1, \gamma=0.1$ leads to the best performance of the current task, which means a combination of data and classifier placeholders can jointly improve the model's performance. This also guides the hyper-parameters setup in other tasks.
We also show the influence of classifier placeholder number $C$ in Figure~\ref{fig:supp1b}. Since we arrange 15 classes as the open-set class, we tune it in the range of $\{1,3,\ldots,15\}$. The results validate that multiple dummy classifiers increase the diversity of classifier placeholders, and can match novel patterns with the nearest classifier. However, learning too many dummy classifiers does not help open-set recognition, which shows a decline when $C>11$.

\subsection{Running Time Comparison}

\begin{figure}[t]
	\begin{center}
		\includegraphics[width=.85\columnwidth]{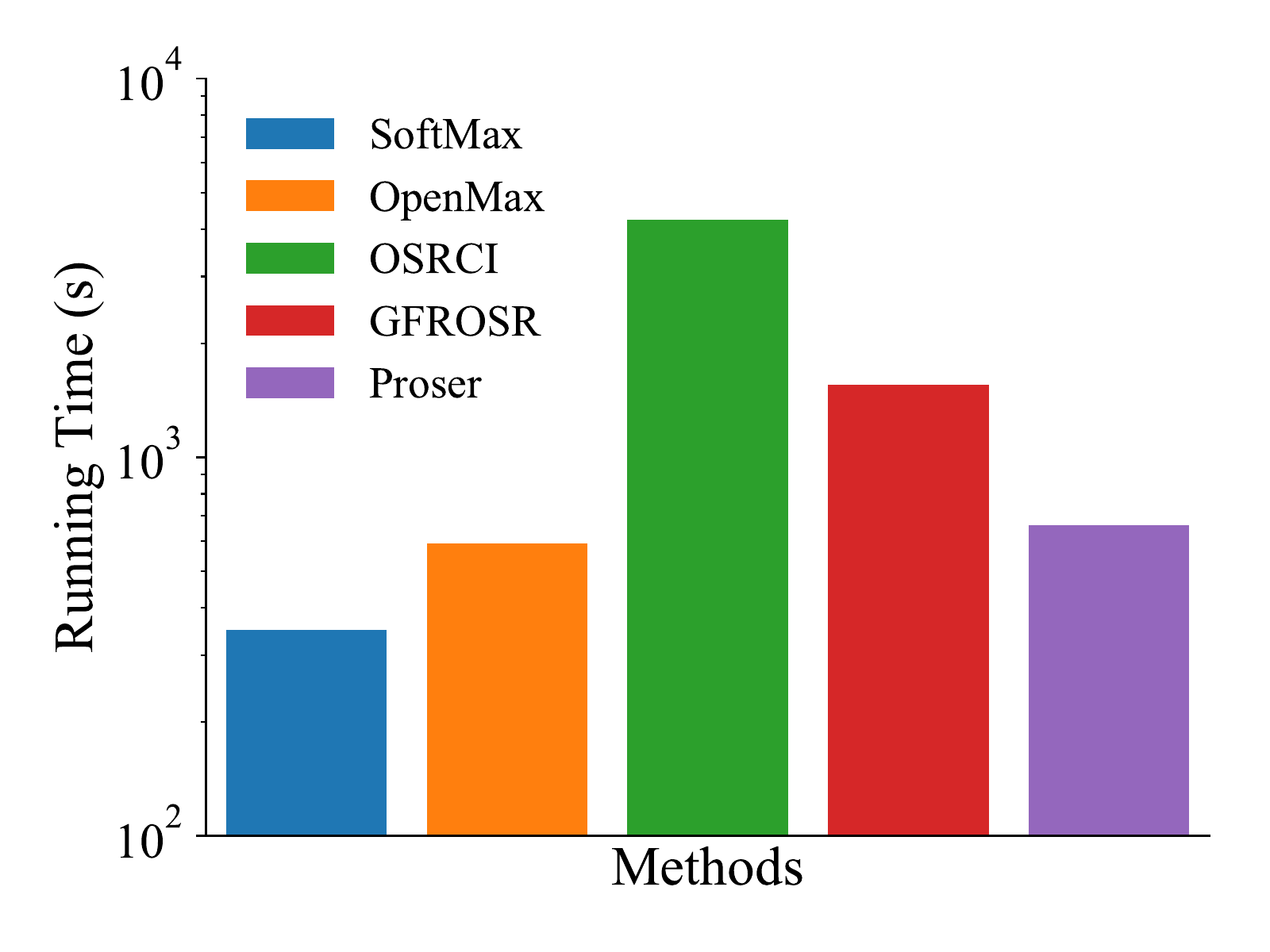}
	\end{center}
	\caption{Running time comparison for different methods on MNIST dataset. The y-axis is in the logarithmic scale.} \label{fig:runtime}
\end{figure}

\firstcite{yoshihashi2019classification} point out that generative-based methods need more time in model training and instance generating. As a result, it takes more time to implement these methods in real-world applications. We conduct experiments on MNIST, and compare to \cite{neal2018open,perera2020generative,bendale2016towards} as well as softmax in terms of the training time. The results are reported in Figure~\ref{fig:runtime}. The running time of  generative methods includes training generative models and novel instance generation. 

From Figure~\ref{fig:runtime}, we can tell that \name is of the same order of magnitude with Softmax and OpenMax~\cite{bendale2016towards}. In comparison, generative-based methods OSRCI~\cite{neal2018open} and GFROSR~\cite{perera2020generative}  consume much more time than \mame. OSRCI needs to generate counterfactual images as novelty, and augment the initial dataset with these generated open-set instances, which consumes the most time. GFROSR trains an extra generative model to reconstruct images, and the reconstructed images are fed into the classification model, which consumes the second-most time.  Since we only generate open-set classes with manifold mixup, the mixup loss is based on mixed embedding, which means we do not need more steps in training novel patterns. As a result, \name can generate novel classes and train models efficiently.

\subsection{Manifold Mixup VS Vanilla Mixup}

In the main paper, we discuss the pros and cons of manifold mixup~\cite{verma2019manifold} and the reason we do not adopt vanilla mixup in the input space. In this section, we give the detailed pseudo code for manifold mixup for data placeholders and the performance comparison between manifold mixup and vanilla mixup~\cite{zhang2018mixup,tokozume2018between}. 

The guideline of generating data placeholders with manifold mixup is shown in Algorithm~\ref{supp:alg1}. Comparing to vanilla training, where mini-batch instances are fed into the model to forward passing and conduct back-propagation, our proposed method consumes the same complexity of forwarding and backward passing. Line~\ref{line1} forward the mini-batch with the pre-embedding module, and get the middle representation of the original batch. These middle-representations are then shuffled with random order to form $\mathcal{D}_{\text{shuffle}}$, as shown in Line~\ref{line2}. To avoid mixing two instances from the same class, we mask the pairs of the same class with Line~\ref{line4}, and then conduct mixup to generate data placeholders in Line~\ref{line7}. 

Note that the size of mixed embeddings $\tilde{\x}_{pre}$ should be no more than $B$, since we only combine unmasked instances. These data placeholders are then fed into the post-embedding module to get the ultimate representation in Line~\ref{line8}. As a result, the total forward and backward consumption is no more than vanilla training.

\begin{algorithm}[t]
	\caption{ Manifold mixup for data placeholders }
	\label{supp:alg1}
	{\bf Input}: Embedding module  $\phi(\cdot)$, which can be decomposed of pre-embedding  $\phi_{pre}(\cdot)$ and	post-embedding  $\phi_{post}(\cdot)$;\\
	Closed-set mini-batch: $\mathcal{D}_{tr}= \left\{ \left( \x _ { i }  , y _ { i }  \right) \right\} _ { i = 1 } ^ { B }$;

	{\bf Output}:	Updated  classifier $\hat{f}$; 
	\begin{algorithmic}[1]{
			\STATE Calculate the pre-embeddings of this mini-batch $\phi_{pre}(\mathcal{D}_{tr})$;\label{line1}
			\STATE Shuffle the mini-batch with random order, and get shuffled (embedding,label) pair $\mathcal{D}_{\text{shuffle}}= \left\{ \left(\phi_{pre}( \hat{\x} _ { i } ) , \hat{y} _ { i }  \right) \right\} _ { i = 1 } ^ { B }$; \label{line2}
			\FOR{$i=1,\cdots,B$}
			\STATE Mask the pairs of the same class, \ie,  $y _ { i }=\hat{y} _ { i }$ \label{line4};
			\ENDFOR
			\STATE Sample $\lambda$ from Beta distribution;
			\STATE Calculate the manifold mixup pre-embeddings $\tilde{\x}_{pre}$ with unmasked pairs, \ie, data placeholders; \label{line7}
			\STATE Calculate the post-embeddings of $\tilde{\x}_{pre}$, \ie, $\phi_{post}(\tilde{\x}_{pre})$; \label{line8}
			\STATE Calculate the manifold mixup loss $\leftarrow$ Eq. 7;
		}
	\end{algorithmic}
	
\end{algorithm}
\begin{figure}[t]
	\begin{center}
		\subfigure[Vanilla mixup]
		{\includegraphics[width=.485\columnwidth]{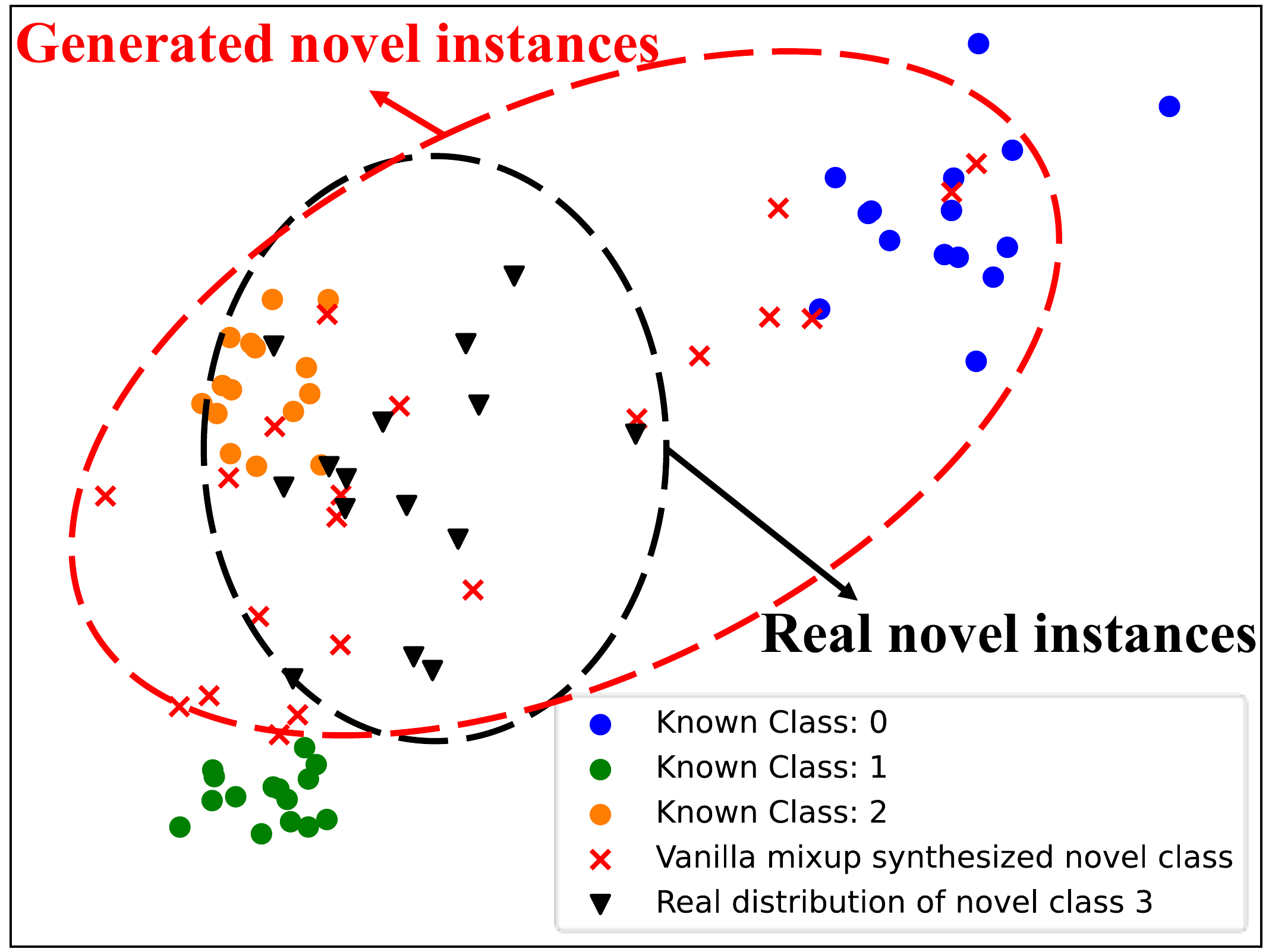}\label{fig:supp4a}}
		\subfigure[Manifold mixup]
		{\includegraphics[width=.495\columnwidth]{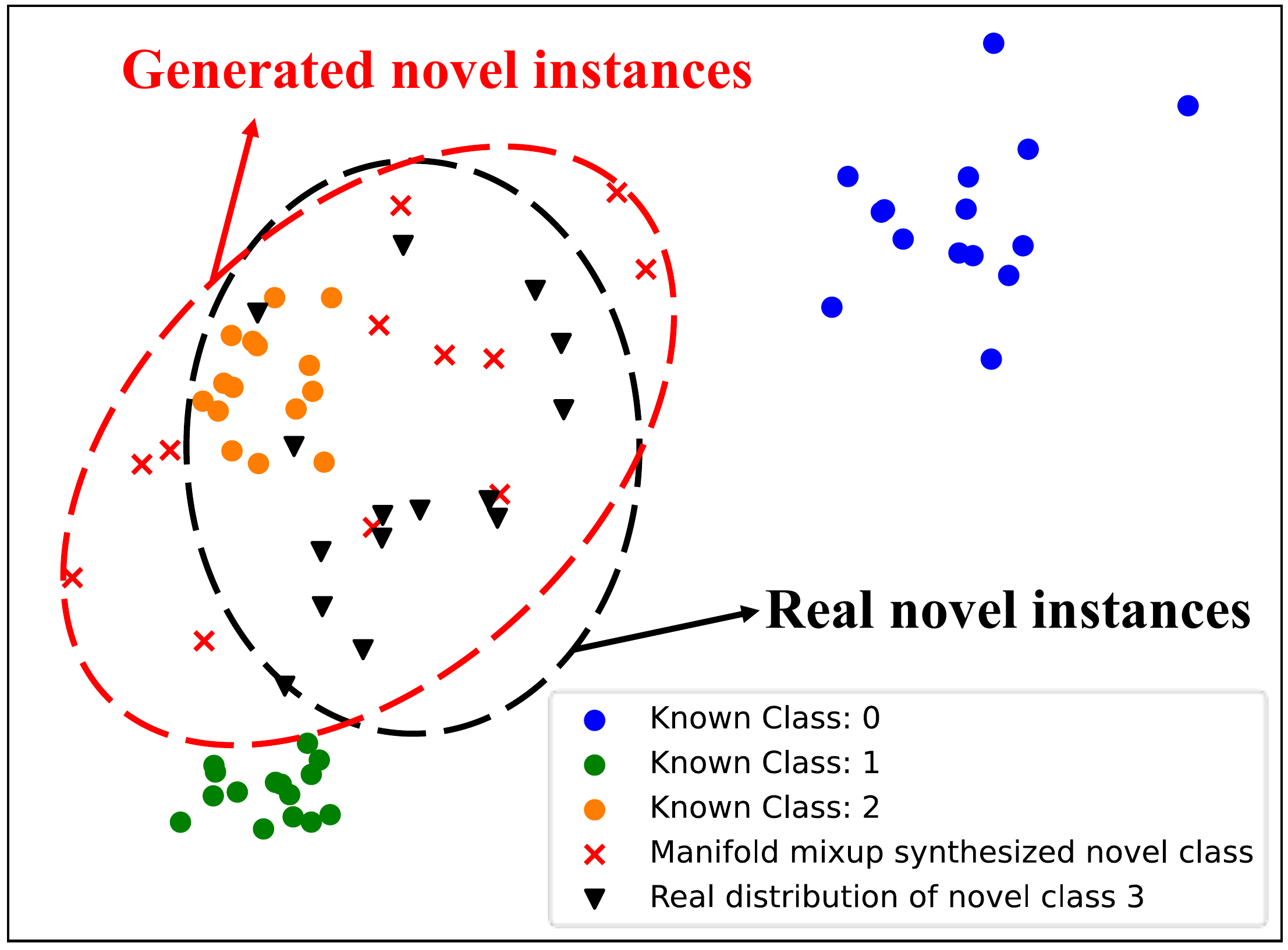}\label{fig:supp4b}}
		
	\end{center}
	\caption{Visualization of generated novel instances and real distribution of novel instances. Left: novel instances generated by vanilla mixup. Right: novel instances generated by manifold mixup. Red  crosses stand for generated novel instances, colored dots stand for known class instances, and black crosses stand for real distribution of novel class. The generated space of vanilla mixup covers known space, which may hurt the learning of embedding space.} \label{fig:supptsne}
\end{figure}
\begin{table}[t] 
	\centering{
		\caption{Ablation study over manifold mixup and vanilla mixup, the configurations are the same as in Figure~\ref{fig:suppparam}.}
		{\begin{tabular}{l|c|c|c|c}
				\addlinespace
				\toprule
				{$\alpha$} &{0.4}  &1 & 1.5 &  2\\
				\midrule
				Vanilla Mixup & 62.3  &  64.4 & 63.9 &64.1  \\	
				\midrule
				Manifold Mixup  & \bf63.0 & \bf68.6 &\bf 70.0 & \bf70.7\\
				\bottomrule
			\end{tabular}\label{table:suppablation}
		}
	}
\end{table}

\begin{figure*}[t]
	\begin{center}
		\subfigure[$\alpha=0.4$]
		{\includegraphics[width=.66\columnwidth]{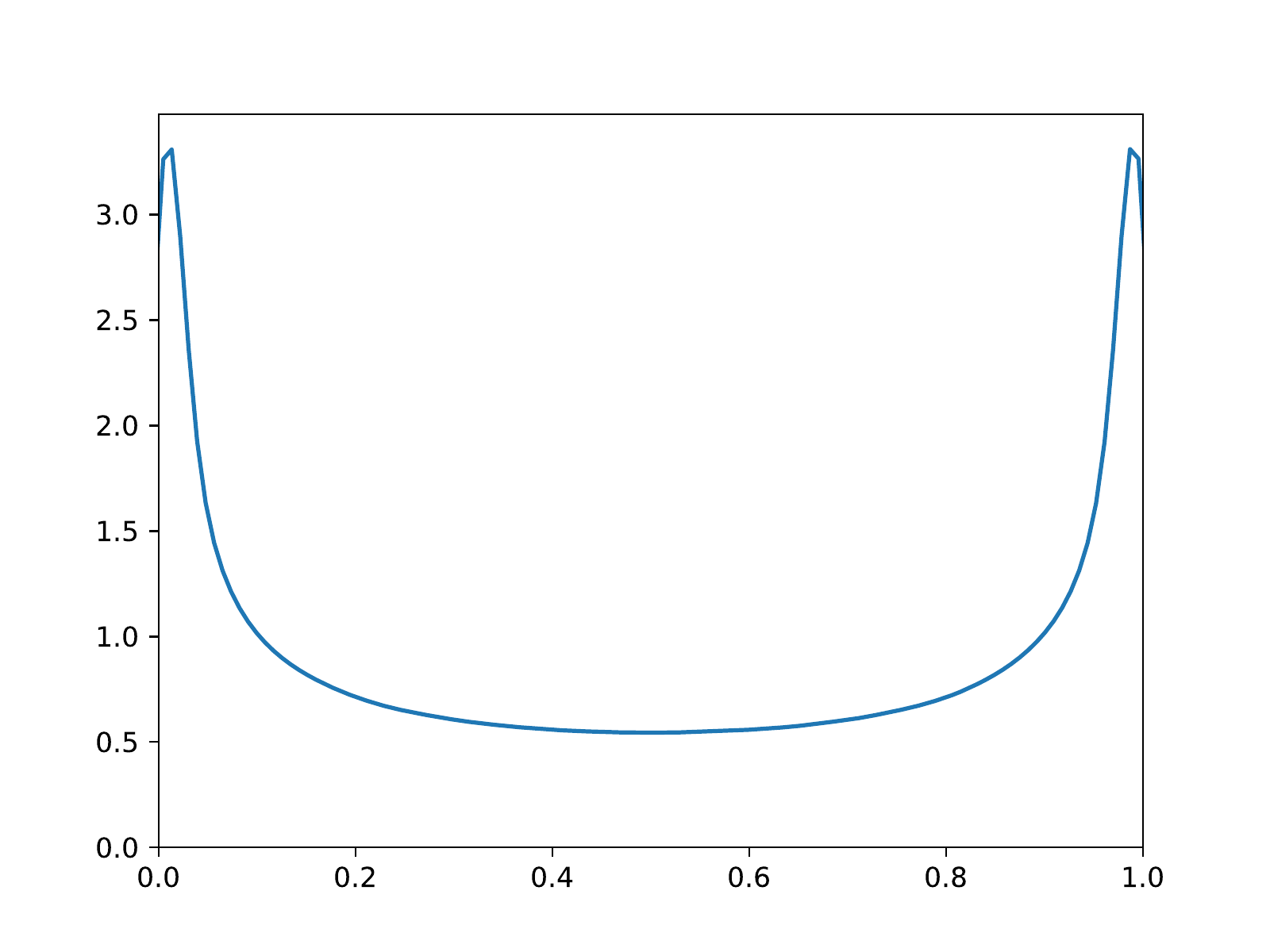}\label{fig:supp3a}}
		\subfigure[$\alpha=1$]
		{\includegraphics[width=.66\columnwidth]{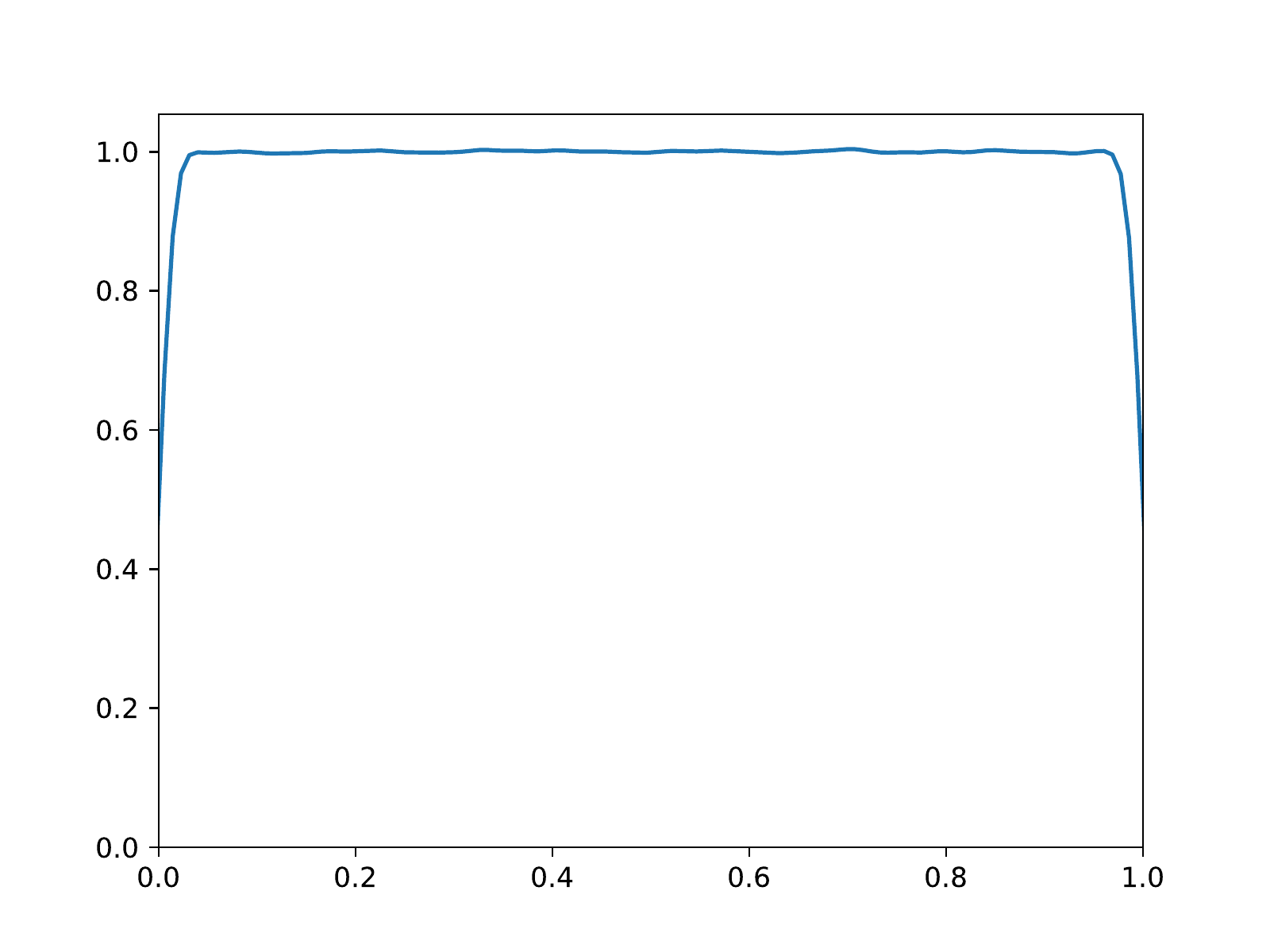}\label{fig:supp3b}}
		\subfigure[$\alpha=2$]
		{\includegraphics[width=.66\columnwidth]{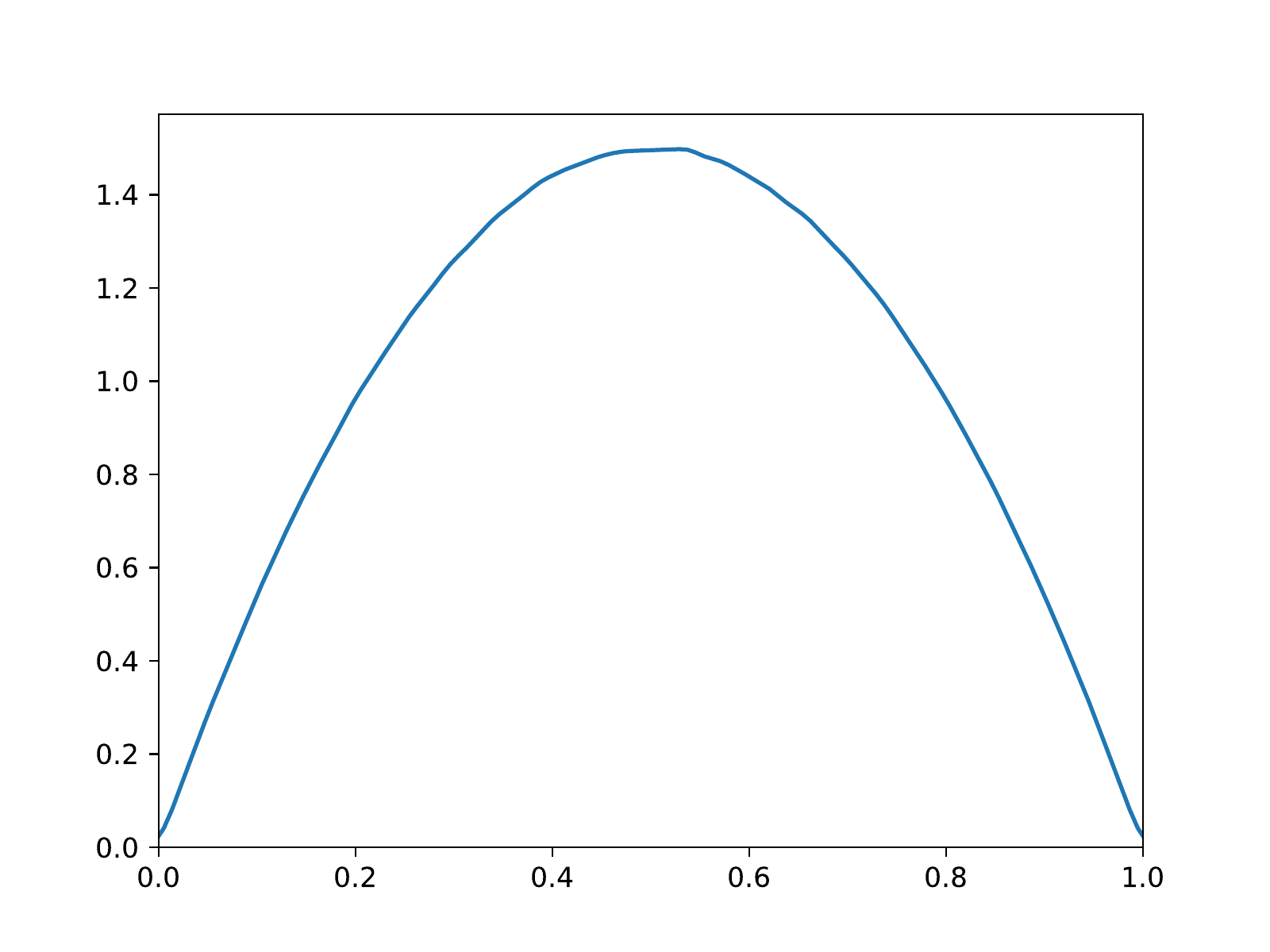}\label{fig:supp3c}}
	\end{center}
	\vspace{-4mm}
	\caption{Kernel density estimation of Beta distribution when $\alpha$ varies.} \label{fig:suppbeta}
	
\end{figure*}

\begin{table*}[t] 
	\centering{
		\caption{Unknown detection performance in terms of AUC (mean$\pm$std). Results are averaged among five randomized trials. N. R. means the original work did not provide a particular value.}
		
		{\begin{tabular}{l|c|c|c|c|c}
				\addlinespace
				\toprule
				{Methods} &{SVHN}  &{CIFAR10}  &{CIFAR+10}  &{CIFAR+50}    
				& Tiny-ImageNet \\
				\midrule
				Softmax &  88.6 $\pm$ 1.4& 67.7 $\pm$ 3.8   & 81.6 $\pm$ \nrr& 80.5 $\pm$ \nrr   &57.7 $\pm$ \nrr \\
				
				OpenMax~\cite{bendale2016towards} & 89.4 $\pm$ 1.3&  69.5 $\pm$ 4.4   & 81.7 $\pm$ \nrr& 79.6 $\pm$ \nrr   &57.6 $\pm$ \nrr\\
				
				G-OpenMax~\cite{ge2017generative} & 89.6 $\pm$ 1.7&  67.5 $\pm$ 4.4  & 82.7 $\pm$ \nrr& 81.9 $\pm$ \nrr   &58.0 $\pm$ \nrr\\
				%(ECCV18)
				OSRCI~\cite{neal2018open} & 91.0 $\pm$ 1.0&  69.9 $\pm$ 3.8   & 83.8 $\pm$ \nrr& 82.7 $\pm$ \nrr   &58.6 $\pm$ \nrr\\
				
				C2AE~\cite{oza2019c2ae}& 89.2 $\pm$ 1.3&  71.1 $\pm$ 0.8   & 81.0 $\pm$ 0.5& 80.3 $\pm$ 0.0   &58.1 $\pm$ 1.9\\
				
				CROSR~\cite{yoshihashi2019classification} & 89.9 $\pm$ 1.8&  \nrr   & \nrr&  \nrr   &58.9 $\pm$ \nrr\\
				
				GFROSR~\cite{perera2020generative} & 93.5 $\pm$ 1.8&  83.1 $\pm$ 3.9   & 91.5 $\pm$ 0.2& 91.3 $\pm$ 0.2  &64.7 $\pm$ 1.2\\
				
				\midrule
				\name &\bf 94.3 $\pm$ 0.6 &\bf 89.1 $\pm$ 1.6 & \bf 96.0 $\pm$ 0.4& \bf95.3 $\pm$ 0.3&  \bf69.3 $\pm$ 0.5\\
				\bottomrule
			\end{tabular}\label{table:suppunknown}
			\vspace{-3mm}
	}}
\end{table*}
We also test the performance comparison between manifold mixup and vanilla mixup, \ie, replace the $\phi_{pre}$ into identity mapping and mixup in the input space. As we stated in the main paper, we argue that manifold mixup is optimizable, and its benefits help the process of novelty generation. \cite{verma2019manifold} proved that manifold mixup can move the decision boundary away from the data in all directions, which results in a compact embedding space. With the help of these data placeholders, the embeddings of known classes would be much tighter, thus leaving more place for embeddings of unknown classes. Moreover, we conduct ablations to validate the effectiveness of manifold mixup in Table~\ref{table:suppablation}. The experiment configuration is the same as Figure~\ref{fig:suppparam}, and we tune different $\alpha$ in the Beta distribution to choose the best hyper-parameter $\alpha$. Since the mixup percentage $\lambda$ is influenced by $\alpha$, we show the kernel density estimation with different $\alpha$ in Figure~\ref{fig:suppbeta}. We can infer that $\alpha<1$ tends to sample $\lambda$ near 0 or 1, while $\alpha>1$ tends to sample $\lambda$ near 0.5. Correspondingly, $\alpha=1$ would result in a uniform distribution. 

Let us return to our intuition where we try to synthesis novel patterns with known instances. Considering the places beside one class should not be a new class, while the middle point between two classes is often low-confidence area, an intuitive way we seek to mimic novel classes  is to use  $\alpha>1$, as shown in Figure~\ref{fig:supp3c}. To our relief, the results in Table~\ref{table:suppablation} also validate the assumption that $\alpha=2$ leads to the best performance. The results also indicate that compared to vanilla mixup, manifold mixup leads to a more compact embedding space, which boosts  open-set recognition. The results also guide the setup of $\alpha$ in all experiments. We adopt $\alpha=2$ in all experiments in the main paper without tuning the best task-specific value.

We also show the embedding of generated novel instances in Figure~\ref{fig:supptsne}. We conduct experiments under the same setting as the visualization experiment part in the main paper, and show the embedding of generated instances by vanilla mixup and manifold mixup. The $\beta$ parameter is the same between these two methods, and we can infer from Figure~\ref{fig:supptsne} that manifold mixup generates instances more similarly than the vanilla method. Besides, the generated space of vanilla mixup is much more than novel space, which also involves known space. As a result, utilizing vanilla mixup may destroy the learned embedding space to some extent.

\section{Experiment Implementation}
%In this part, we introduce the implementation details, \ie, the full results of unknown detection, hyper-parameter selection, model optimization, and dataset configuration.
\subsection{Unknown Detection Results}

In the main paper, we report the averaged AUC of unknown detection tasks.
We simulate the sampling process over five trials~\cite{neal2018open} to report the mean and standard deviation, and the full results are shown in Table~\ref{table:suppunknown}.
We report the baseline performance from~\cite{perera2020generative,yoshihashi2019classification,neal2018open}. Note that N.R. in the table means that the original paper did not report the standard deviation.
\subsection{Implementation Details of \mame.}

We employ the same backbone architecture as~\cite{perera2020generative,neal2018open}. \name is trained with SGD with momentum of 0.9, and the initial learning rate is set to $0.001$ in the experiment. We fix the batch size to $128$ for all datasets. As we discussed in the hyper-parameter part, we set $\beta=1, \gamma=0.1$, and the number of classifier placeholders is set to $5$ for all datasets. The calibration $bias$ is obtained by ensuring $95\%$ validation data be recognized as known. The $\alpha$ parameter in Beta distribution is set to 2 for all datasets.
We conduct the experiment on Nvidia RTX 2080-Ti GPU with Pytorch 1.6.0.

\subsection{Dataset Configuration.}

\begin{figure}[t]
	\begin{center}
		\includegraphics[width=.95\columnwidth]{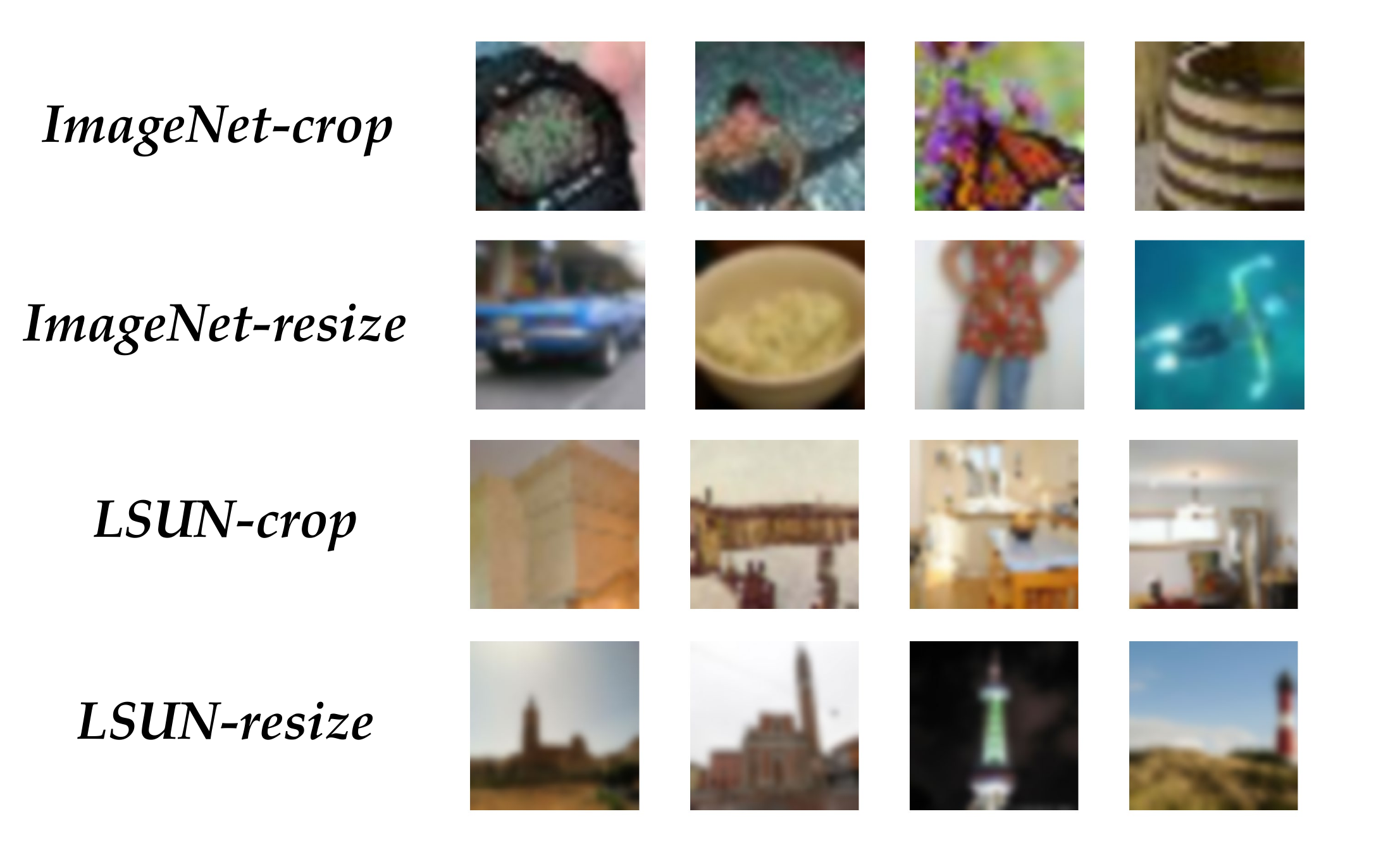}
	\end{center}
	\vspace{-3mm}
	\caption{Dataset example of CIFAR10 open-set recognition, each line stands for the open-set class in main paper.} \label{fig:cifarood}
	\vspace{-3mm}
\end{figure}
We also show the dataset example of CIFAR10 open-set recognition tasks in the main paper in Figure~\ref{fig:cifarood}. The `crop' datasets is part of the original picture, and `resize' datasets is the full original picture resized into 32*32 pixel. As a result, detecting outliers from `crop' datasets is easier than that of `resize' datasets. This is consistent with macro-F1  results in the main paper.

\end{document}